\documentclass{article}

\usepackage[T1]{fontenc}

\usepackage{microtype}
\usepackage{graphicx}
\usepackage{subcaption}
\usepackage{booktabs} 

\usepackage{hyperref}

\PassOptionsToPackage{noend}{algorithmic}
\usepackage[preprint]{icml2026}
\renewcommand{\algorithmicrequire}{\textbf{Input:}}
\renewcommand{\algorithmicensure}{\textbf{Output:}}

\usepackage{amsmath}
\usepackage{amssymb}
\usepackage{mathtools}
\usepackage{amsthm}

\usepackage[capitalize,noabbrev]{cleveref}

\theoremstyle{plain}

\theoremstyle{definition}

\theoremstyle{remark}

\usepackage[disable,textsize=tiny]{todonotes}

\icmltitlerunning{HTN-Orchestrate}

\begin{document}

\twocolumn[
  \icmltitle{HTN-Orchestrate: A Hierarchical Task Network Framework for Multi-Agent Orchestration}



  \icmlsetsymbol{equal}{*}

  \begin{icmlauthorlist}
    \icmlauthor{Yanchen Deng}{ntu}
    \icmlauthor{Jianguo Huang}{ntu}
    \icmlauthor{Yao Long Teng}{ntu}
    \icmlauthor{Yixuan Li}{ntu}
    \icmlauthor{Handuo Zhang}{apodex}
    \icmlauthor{Ning Wang}{apodex}
    \icmlauthor{Liangcai Su}{apodex}
    \icmlauthor{Xinyu Wang}{apodex}
    \icmlauthor{Bo An}{ntu}
  \end{icmlauthorlist}

  \icmlaffiliation{ntu}{Nanyang Technological University, Singapore}
  \icmlaffiliation{apodex}{Apodex}

  \icmlcorrespondingauthor{Bo An}{boan@ntu.edu.sg}

  \icmlkeywords{Machine Learning, ICML}

  \vskip 0.3in
]



\printAffiliationsAndNotice{This work was conducted during the main contributors' internships at Apodex.}

\begin{abstract}
Orchestrating multiple agents on a complex task requires reconciling local subtask execution with global task planning. Dynamic dispatch approaches provide execution flexibility but do not explicitly model subtask hierarchies or dependencies, while predefined workflows impose this structure but fix the decomposition before execution begins. Therefore, we introduce HTN-Orchestrate, a two-layer framework that separates these responsibilities through a hierarchical task network (HTN). The Planner maintains subtask hierarchies and dependencies, deferring decomposition until prerequisite results become available, while local Orchestrators schedule workers to execute each leaf subtask. Results propagate upward under \textbf{AND} semantics, and accumulated findings inform a rebuilt task network when the current decomposition cannot resolve the task. On five benchmarks spanning web research, scientific reasoning, and research workflows, HTN-Orchestrate outperforms ReAct, Agent Team, and DeepSeek Harness using the same base model, achieving 89.73\% accuracy on BrowseComp and 73.44\% on SUPERChem, exceeding Agent Team by 2.45 and 4.69 percentage points, respectively.

\end{abstract}

\section{Introduction}
\label{sec:introduction}
Large language models (LLMs) have demonstrated strong performance on challenging reasoning tasks~\citep{yang2023leandojo,hubert2026olympiad,novikov2025alphaevolve}, and by integrating with external tools, LLM-powered agents~\citep{yao2023react,wang2025openhands,agashe2025agents2} can proactively acquire information and revise actions in response to observations. Multi-agent systems~\citep{wu2024autogen,openclaw2026,hermesagent2026,claudecode2025,claudemanagedagents2026} further extend this approach by organizing agents into complementary roles that distribute investigations, explore independent approaches, and verify candidate solutions. Directing these activities toward a common goal requires \emph{orchestration}~\citep{zhang2025agentorchestra,ruan2026aorchestra}, which must reconcile local subtask execution that responds to emerging findings with global task planning that maintains a structured decomposition and tracks dependencies across subtasks.

\begin{figure*}[t]
\centering
\includegraphics[width=\textwidth]{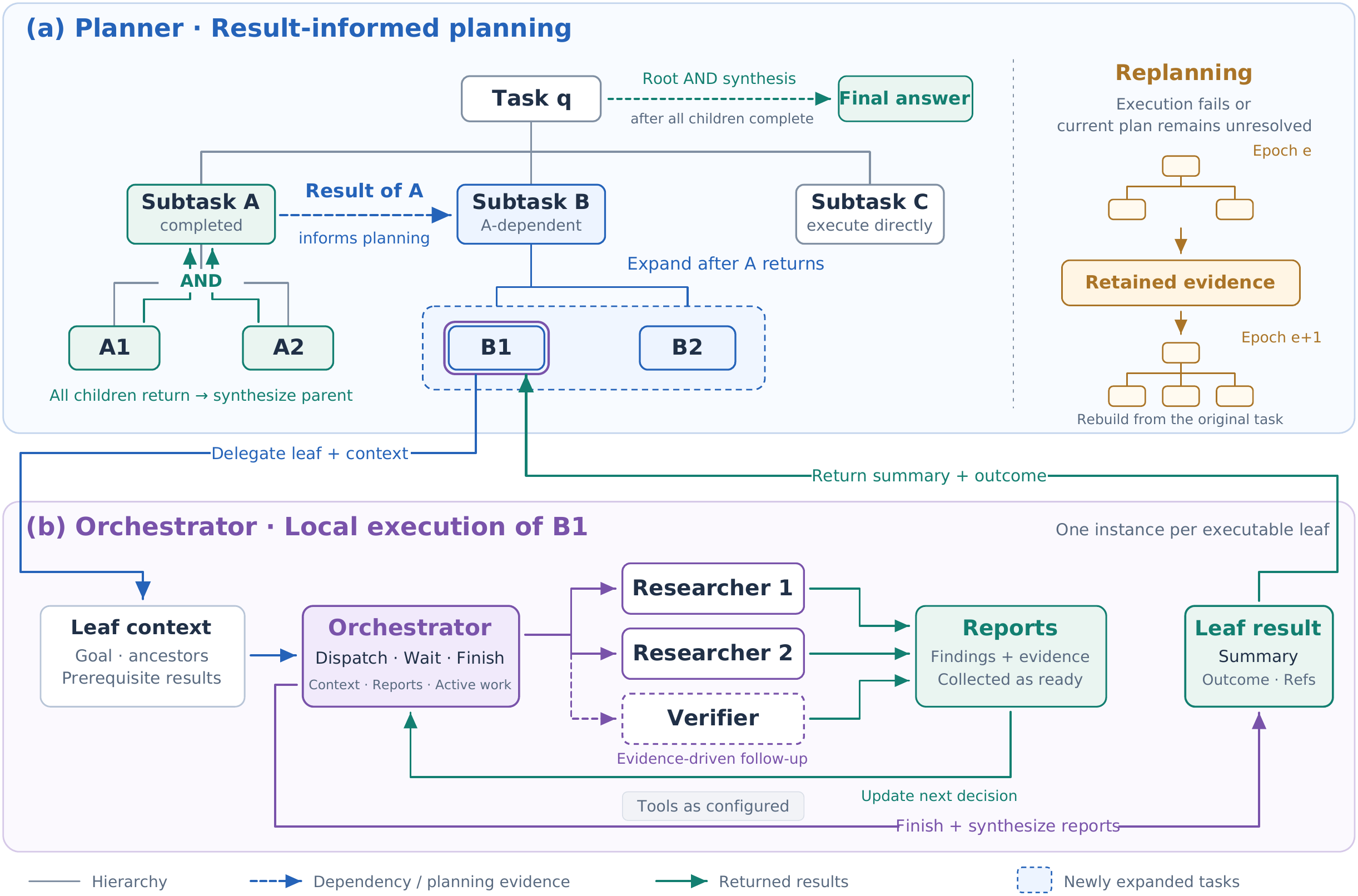}
\caption{HTN-Orchestrate separates result-informed task planning from local multi-agent orchestration. (a) The Planner expands ready tasks using prerequisite results and synthesizes parent results after all children return. When rebuilding is needed, supported evidence is retained to construct a new task network. (b) Each executable leaf is delegated to an Orchestrator, which dispatches workers, collects reports as they become available, and uses those reports to guide subsequent assignments and leaf synthesis. The leaf result returns to the task network for further planning and upward synthesis.}
\label{fig:htn-orchestrate-overview}
\end{figure*}

Existing approaches address orchestration through structured workflows or runtime delegation, but each sacrifices one side of this tradeoff. Structured workflows organize tasks into stages connected by explicit control flow, specifying when agents act and how they exchange results~\citep{hong2024metagpt,langgraph2026workflows}. Whether authored manually or generated automatically as in AFlow~\citep{zhang2025aflow}, these workflows impose clear global structure but fix the decomposition before execution begins, so that findings from local execution cannot reshape subsequent planning. Runtime delegation addresses this rigidity by having a central orchestrator determine subtasks and assign agents as execution proceeds~\citep{fourney2024magentic,ruan2026aorchestra}, with agent-team approaches~\citep{kimi2026k25,apodex2026apodex11} further extending this to parallel subagent dispatch. However, this design couples task decomposition, subtask dispatch, and result synthesis within a single orchestrator~\citep{ruan2026aorchestra,kimi2026k25,zhang2025agentorchestra}, without explicitly modeling subtask hierarchies or precedence dependencies among sibling subtasks. Although dynamic decomposition methods such as ADaPT~\citep{prasad2024adapt} and ReAcTree~\citep{choi2026reactree} allow the plan to evolve during execution, they too provide limited support for representing complex dependencies, leaving the tension between adaptive local execution and structured global planning unresolved.

We introduce \textbf{HTN-Orchestrate}\footnote{The implementation of HTN-Orchestrate is available at \url{https://github.com/ycdeng-ntu/htn-orchestrate}}, a two-layer orchestration framework that separates these responsibilities through a hierarchical task network (HTN)~\citep{erol1994htn}. At the upper layer, the Planner expands the task network incrementally as prerequisite results become available, so that subsequent decomposition decisions reflect information obtained during execution. At the lower layer, each executable leaf subtask is delegated to an Orchestrator that schedules workers (e.g., researchers, verifiers) and uses their reports to guide investigation, verification, and synthesis. Results propagate upward through the task network under \textbf{AND} semantics, whereby each leaf subtask returns the result produced by its Orchestrator, while each non-leaf subtask synthesizes its result after all its children have completed. The framework also incorporates a rebuilding mechanism that constructs a new task network conditioned on partial results from the previous epoch when execution fails or there is no plausible solution under the current task network.

Our main contributions are as follows.
\begin{itemize}
  \item We propose HTN-Orchestrate, a two-layer orchestration framework (Figure~\ref{fig:htn-orchestrate-overview}) that decouples global task planning from local subtask execution through a hierarchical task network that explicitly represents subtask hierarchies and precedence dependencies. The Planner incrementally expands the network as prerequisite results become available, while each executable leaf subtask is delegated to an Orchestrator that coordinates multiple workers.
  \item We combine result-informed planning, which defers decomposition decisions for dependent subtasks until prerequisite findings are available, with AND-semantic synthesis that propagates results upward through the task hierarchy, and a rebuilding mechanism that retains supported evidence from completed work to construct a revised task network when the current decomposition cannot resolve the task.
  \item We evaluate HTN-Orchestrate on five benchmarks spanning web research, scientific reasoning, and research workflows and demonstrate that it consistently outperforms Agent Team~\citep{apodex2026apodex11} and DeepSeek Harness~\citep{deepseek-harness2026}. In particular, HTN-Orchestrate achieves 89.73\% accuracy on BrowseComp and 73.44\% on SUPERChem, exceeding Agent Team by 2.45 and 4.69 percentage points, respectively. Ablations on BrowseComp demonstrate the contributions of result-informed planning, local multi-agent orchestration, and task-network rebuilding.
\end{itemize}

\section{Related Work}
\label{sec:related-work}
\subsection{Multi-Agent Orchestration}
\label{sec:related-orchestration}

Multi-agent orchestration methods vary in how they distribute responsibilities between global task planning and local execution, and most designs favor one side of this tradeoff. Early frameworks such as AutoGen~\citep{wu2024autogen} and MetaGPT~\citep{hong2024metagpt} organize collaboration through conversation patterns and role-based procedures that provide clear global structure but follow fixed interaction protocols. AFlow~\citep{zhang2025aflow} and MaAS~\citep{zhang2025maas} reduce manual design effort by automating workflow and architecture construction, while Conductor~\citep{nielsen2026conductor} and MAS-Orchestra~\citep{ke2026masorchestra} learn to generate orchestration decisions through reinforcement learning. Although these methods can discover effective coordination patterns, the resulting structures are determined before or independently of task execution.

Runtime delegation shifts toward local adaptability by allowing an orchestrator to assign work as execution proceeds, with Magentic-One~\citep{fourney2024magentic} and AOrchestra~\citep{ruan2026aorchestra} managing task progress through agent feedback. Kimi K2.5~\citep{kimi2026k25} and Apodex 1.1~\citep{apodex2026apodex11} further support parallel subagents dynamically dispatched by a main agent, and such dispatch decisions can also be learned through policies for agent selection and tool invocation~\citep{dang2025evolving,su2026toolorchestra}. However, these approaches couple task decomposition, subtask dispatch, and result synthesis within a single orchestrator, without explicitly separating global planning from local execution management. Workforce, introduced with OWL~\citep{hu2025owl}, addresses this by separating a domain-agnostic planner from a central coordinator that manages worker assignments and task dependencies. HTN-Orchestrate shares this separation but assigns a dedicated Orchestrator to each executable leaf subtask, while the Planner maintains and incrementally expands the task network, allowing local coordination within each subtask to operate independently of global planning over the task hierarchy.

\subsection{Task Decomposition}
\label{sec:related-decomposition}

The tension between local execution and global planning is particularly evident in task decomposition, where the challenge is determining not only how to break a task apart but also when to revise that breakdown in response to execution findings. Static approaches commit to a full decomposition before observing results. Least-to-most prompting~\citep{zhou2023leasttomost} generates subproblems before solving them sequentially, and Agent-Oriented Planning~\citep{li2025aop} accounts for agent capabilities when decomposing and allocating tasks. D-CORE~\citep{xu2026dcore} uses reinforcement learning to incentivize decomposition reasoning in large reasoning models, and Select-Then-Decompose~\citep{liu2025selectdecompose} chooses a decomposition strategy according to the task. These methods improve the quality of decomposition but still finalize it before execution begins.

When intermediate findings affect subsequent work, decomposition must also adapt during execution. Plan-and-Act~\citep{erdogan2025planandact} trains models on planning trajectories and revises the plan after each executor step, ADaPT~\citep{prasad2024adapt} recursively decomposes tasks that resist direct execution, ReCAP~\citep{zhang2025recap} revises the remaining plan using execution results, and ReAcTree~\citep{choi2026reactree} organizes execution through hierarchical agent trees. TodoEvolve~\citep{liu2026todoevolve} extends adaptation to the planning architecture itself by learning to generate and revise task-specific planning systems. These methods allow plans to evolve but provide limited support for explicitly modeling precedence dependencies among subtasks. HTN-Orchestrate uses prerequisite results to inform the further decomposition of dependent subtasks, deferring decisions about their internal structure until the relevant evidence becomes available.

\subsection{Hierarchical Task Networks and Dependency-Aware Planning}
\label{sec:related-htn}

Explicit dependencies connect global planning with local execution by determining which results must be available before subsequent work can proceed. Hierarchical task networks represent complex activities through task decomposition and constraints on the resulting subtasks~\citep{erol1994htn}, and recent work such as ChatHTN~\citep{munozavila2025chathtn} combines this tradition with language models by interleaving LLM-based decomposition with symbolic HTN planning. At a different level of granularity, LLMCompiler~\citep{kim2024llmcompiler} constructs dependencies among function calls to schedule parallel execution, whereas FlowSearch~\citep{hu2025flowsearch} models research as an evolving graph of interdependent knowledge units and dynamically revises its structure in response to intermediate findings. HTN-Orchestrate distinguishes the parent-child relation that defines a decomposition from precedence dependencies among sibling subtasks. The former determines how results are synthesized upward after all children complete, while the latter determines which prerequisite results are available for further decomposition and execution. Each executable task then invokes an Orchestrator that can organize multiple workers, combining hierarchical decomposition, result-informed planning, and local multi-agent coordination within a common execution process. Table~\ref{tab:feature-comparison} summarizes the feature differences between HTN-Orchestrate and the most closely related systems.

\begin{table}[t]
\caption{Feature comparison with closely related systems. STR = structured plan (explicit task hierarchy or dependency graph); RES = result-informed (subsequent work conditioned on execution findings); DEC = decoupled (global planning and local execution handled by separate controllers); RPL = replanning (task structure revised during execution); EVC = evidence carryover (findings from a prior plan explicitly condition the rebuilt plan).}
\label{tab:feature-comparison}
\centering
\resizebox{\columnwidth}{!}{%
\begin{tabular}{@{}lccccc@{}}
\toprule
& STR & RES & DEC & RPL & EVC \\
\midrule
Agent Team~\citep{apodex2026apodex11} & & $\checkmark$ & & & \\
Workforce~\citep{hu2025owl} & $\checkmark$ & & $\checkmark$ & $\checkmark$ & \\
ADaPT~\citep{prasad2024adapt} & $\checkmark$ & & & $\checkmark$ & \\
FlowSearch~\citep{hu2025flowsearch} & $\checkmark$ & $\checkmark$ & & $\checkmark$ & \\
ChatHTN~\citep{munozavila2025chathtn} & $\checkmark$ & & & & \\
HTN-Orchestrate (ours) & $\checkmark$ & $\checkmark$ & $\checkmark$ & $\checkmark$ & $\checkmark$ \\
\bottomrule
\end{tabular}}
\end{table}

\section{Method}
\label{sec:method}
Given a task $q$ and a set of worker roles and tools, HTN-Orchestrate separates global task planning from local subtask execution through two cooperating layers (Figure~\ref{fig:htn-orchestrate-overview}). At the upper layer, the Planner incrementally constructs a hierarchical task network that encodes subtask decompositions and their dependencies, while at the lower layer, each executable leaf subtask is delegated to an Orchestrator that schedules workers and returns a result. Because these results flow back to the Planner, subsequent decomposition decisions reflect findings that were unavailable when the network was first constructed, and when the current decomposition cannot resolve the task, accumulated evidence supports the construction of a revised network.

\begin{algorithm}[t]
\caption{Planner: Result-Informed Planning and Replanning}
\label{alg:planner}
\begin{algorithmic}[1]
\REQUIRE Task $q$, workers, rebuild limit $B$, execution budgets
\ENSURE Answer and resolution outcome
\STATE $M\gets\varnothing$, epoch $e\gets 0$
\LOOP
  \STATE $\mathcal{T}\gets\mathrm{InitializeRoot}(q,M)$
  \WHILE{epoch can advance within execution budgets}
    \WHILE{a ready planning leaf $v$ exists}
      \STATE $\mathrm{PlanOrExecute}(v,\mathcal{T},M)$
    \ENDWHILE
    \STATE $L\gets$ ready executable leaves of $\mathcal{T}$
    \IF{$L=\varnothing$}
      \STATE \textbf{break}
    \ENDIF
    \STATE Start $\mathrm{Orchestrate}(v,C_v)$ for all $v\in L$
    \STATE As results arrive, $\mathrm{PropagateResults}(\mathcal{T})$
    \STATE Settle the wave, or stop it for replanning
    \IF{the root is resolved}
      \STATE \textbf{return} root result and outcome
    \ENDIF
  \ENDWHILE
  \IF{$e\geq B$ or execution budgets are exhausted}
    \STATE \textbf{return} synthesis of available evidence
  \ENDIF
  \STATE $M\gets\mathrm{CollectEvidence}(\mathcal{T},M)$
  \STATE $e\gets e+1$
\ENDLOOP
\end{algorithmic}
\end{algorithm}

\begin{algorithm}[t]
\caption{Orchestrator: Multi-Agent Subtask Execution}
\label{alg:orchestrator}
\begin{algorithmic}[1]
\REQUIRE Leaf $v$, context $C_v$, worker roles, budgets
\ENSURE Result summary, outcome, and evidence references
\STATE $R_v\gets\varnothing$, $A_v\gets\varnothing$
\WHILE{dispatch-round and time budgets remain}
  \STATE Collect finished reports into $R_v$ and update $A_v$
  \STATE $a\gets\mathrm{Decide}(C_v,R_v,A_v)$
  \IF{$a$ is dispatch}
    \STATE Select roles and generate worker instructions
    \STATE Add assignments to $A_v$ within available limits
    \STATE Collect progress within the allotted waiting time
  \ELSIF{$a$ is wait}
    \STATE Wait for worker progress within the time budget
  \ELSE
    \STATE Stop unnecessary active assignments
    \STATE \textbf{return} $\mathrm{SynthesizeLeaf}(C_v,R_v)$
  \ENDIF
  \IF{the shared worker-dispatch budget is exhausted}
    \STATE \textbf{break}
  \ENDIF
\ENDWHILE
\STATE Collect available reports within remaining limits
\STATE Stop remaining assignments and update $R_v$
\STATE \textbf{return} $\mathrm{SynthesizeLeaf}(C_v,R_v)$
\end{algorithmic}
\end{algorithm}

\subsection{Task Network Representation}
\label{sec:method-representation}

Following the HTN formulation~\citep{erol1994htn}, we represent a task network as $\mathcal{T}=(V,E_h,E_d)$, whose root corresponds to $q$. The hierarchy edges $E_h$ form a rooted tree in which each parent is decomposed into children that jointly address its task, and the dependency edges $E_d$ impose acyclic precedence constraints among siblings, with an edge $(u,v)$ indicating that $v$ requires the result of $u$. We denote the children and direct prerequisites of $v$ by $\mathrm{Ch}(v)$ and $\mathrm{Pred}(v)$, respectively. Unlike classical HTN planners that draw from a predefined library of decomposition rules, our Planner uses an LLM to generate decompositions during execution, which allows the network to grow incrementally as prerequisite results become available.

Each node stores its task description, execution status, and result summary $y_v$ (Appendix~\ref{app:implementation} lists the full node fields and prompt templates). A node with $\mathrm{Ch}(v)=\varnothing$ is a leaf, which the Planner may further decompose into children or route to an Orchestrator for direct execution. Before either can happen, the leaf must be ready, meaning it is pending and all its prerequisites have completed,
\begin{equation}
\begin{aligned}
\mathrm{Ready}(v) \iff {}& \mathrm{Ch}(v)=\varnothing\ \land\ v\text{ is pending} \\
 &{}\land\ \forall u\in\mathrm{Pred}(v),\ u\text{ is completed}.
\end{aligned}
\label{eq:task-readiness}
\end{equation}
Because execution can yield partial findings rather than a definitive answer, each result also records whether the subtask is resolved or remains open. The completed status in the Ready condition encompasses both outcomes, so a downstream task proceeds with whatever evidence its prerequisite produced regardless of resolution. The resolved-or-open distinction instead feeds into synthesis and replanning decisions that determine whether further work is needed.

\subsection{Result-Informed Planning}
\label{sec:method-planning}

The Planner uses prerequisite execution results to guide subsequent task decomposition and retains accumulated findings to support replanning, as summarized in Algorithm~\ref{alg:planner}. For each ready leaf, $\mathrm{PlanOrExecute}$ considers the original goal, the task description, its ancestors, and summaries of relevant prerequisite results, and based on this context either marks the task for direct execution or expands the network with children and their sibling dependencies. Because a dependent subtask receives its decomposition only after its prerequisites return, its internal structure can reflect their findings rather than relying on assumptions made before execution began. For example, a task that first identifies a research subject can supply its result to a subsequent analysis task, whose decomposition can then target that subject's specific properties (Figure~\ref{fig:case-result-informed} illustrates this on a web-research question where identifying a set of authors allows the Planner to specialize the downstream search to that specific list). Each proposed decomposition is validated for coverage of the parent task and acyclicity of the proposed dependency edges, and revised if either check fails. Depth and branching limits further control overall network growth.

To execute the resulting network, the Planner first processes ready planning nodes until none remain, then dispatches the ready executable leaves to their Orchestrators in parallel. Each such batch forms a wave, and the next planning round begins after the current wave settles. Returned results support both subsequent decomposition along dependency edges and upward synthesis along hierarchy edges. Under AND semantics, an internal node synthesizes its result only after all its children have completed execution. Thus, the result of a node is obtained recursively as
\begin{equation}
 y_v =
 \begin{cases}
 \mathrm{Orchestrate}(v,C_v), & \mathrm{Ch}(v)=\varnothing, \\
 \mathrm{Synthesize}\!\left(v,\{y_u\}_{u\in\mathrm{Ch}(v)}\right),
   & \text{otherwise},
 \end{cases}
\label{eq:result-synthesis}
\end{equation}
where $C_v$ is the execution context defined in Section~\ref{sec:method-orchestration}, and the leaf case applies after the Planner has marked $v$ executable. As leaf results arrive, $\mathrm{PropagateResults}$ records them in the network and recursively synthesizes each parent whose children have all returned their results. Each synthesized summary is passed to the next parent until the root produces the final result for the original task.

When execution fails or the returned findings indicate no plausible solution under the current task network, the Planner can reconsider the decomposition through replanning. Each construction and execution of a task network constitutes an \emph{epoch}, and before starting a new epoch the Planner settles or stops ongoing work and uses $\mathrm{CollectEvidence}$ to extract supported findings from completed and partial reports into an evidence summary $M$. This summary conditions the construction of a new network rooted at the original task, providing planning context rather than automatically marking any new subtask as completed, so that the next epoch can pursue a different decomposition while retaining useful evidence. The number of rebuilds is bounded, and the framework terminates through one of two paths. When the root node is resolved, its synthesized result is returned directly as the final answer. When execution budgets are exhausted before the task is resolved, the framework instead synthesizes the evidence obtained so far and reports any unresolved requirements, so the output reflects the sufficiency of the evidence rather than treating the stopping condition as task success.

\subsection{Multi-Agent Orchestration}
\label{sec:method-orchestration}

For each executable leaf $v$, an Orchestrator manages the workers needed to address that subtask (Algorithm~\ref{alg:orchestrator}). Workers carry out targeted research, reasoning, or verification using their configured tools, while the Orchestrator decides how the investigation should proceed based on their reports. Each Orchestrator receives a stable context $C_v$ containing the original task, the leaf's objective, relevant ancestor descriptions, and prerequisite result summaries, and during execution it maintains the set of returned worker reports $R_v$ and active assignments $A_v$.

At each decision point, the Orchestrator considers $C_v$, $R_v$, and $A_v$ and chooses to dispatch workers, wait for progress, or finish the subtask. A dispatch decision selects worker roles and produces task-specific instructions based on the current evidence, so an initial investigation can be followed by a focused search or a verification step when a report reveals a gap or a disputed claim. Reports are collected as workers finish, allowing the Orchestrator to schedule follow-up work while other assignments remain active, or to wait for new reports when the outstanding questions are already covered. This progressive collection of results allows local decisions to respond to evidence within a wave, while limits on concurrency, dispatch rounds, and execution time bound the work assigned to each subtask.

Once the Orchestrator decides to finish, it stops work that is no longer needed and synthesizes the collected reports into a leaf result that includes a summary, a resolved or unresolved outcome, and evidence references. If an execution limit is reached before a voluntary finish, the Orchestrator collects available reports within the remaining time and returns the supported findings together with any outstanding gaps. In both cases, the outcome reflects the sufficiency of the evidence rather than merely whether workers stopped or a verifier was invoked, and the Planner uses the returned result to inform subsequent decomposition through $\mathrm{PlanOrExecute}$ and parent synthesis through $\mathrm{PropagateResults}$.
\begin{table*}[t]
\caption{Main results. BC, FSO, and SUPERChem report accuracy (\%), FSR reports success rate (\%), and PRL-Bench reports the mean percentage of satisfied rubric criteria. Higher is better for all metrics; bold marks the best score and underline marks the second best on each benchmark.}
\label{tab:main-results}
\centering
\small
\begin{tabular}{lccccc}
\toprule
Method / Model & BC & FSR & FSO & SUPERChem (text-only) & PRL-Bench \\
\midrule
\multicolumn{6}{l}{\textit{Proprietary}} \\
GPT-6 Astra~\citep{openai2026gpt6astra} & \textbf{91.50} & -- & -- & -- & -- \\
GPT-5.6 Sol~\citep{openai2026gpt56} & 90.40 & -- & -- & \textbf{73.60} & -- \\
GPT-5.5~\citep{openai2026gpt55} & 84.40 & 33.90 & 73.80 & 66.20 & -- \\
GPT-5.4~\citep{openai2025gpt5} & 82.70 & 23.33 & -- & 57.70 & 37.38 \\
Claude Fable 5~\citep{anthropic2026fable5} & 87.40 & -- & -- & -- & -- \\
Claude Opus 4.8~\citep{anthropic2026opus48} & 84.30 & 26.70 & 74.30 & 66.90 & -- \\
Claude Opus 4.6~\citep{anthropic2026opus46} & 83.70 & -- & 73.00 & 62.30 & 37.40 \\
Gemini 3.1 Pro~\citep{google2026gemini31} & 85.90 & 16.70 & \textbf{79.00} & 54.30 & \underline{44.27} \\
Qwen 3.7 Max~\citep{qwen2026qwen37} & -- & 24.80 & 74.30 & 60.50 & -- \\
Meta Muse Spark~\citep{meta2026musespark} & -- & 38.00 & -- & -- & -- \\
\midrule
\multicolumn{6}{l}{\textit{Open-weight}} \\
Kimi K3~\citep{kimi2026k3} & \underline{91.20} & 21.70 & 69.00 & 66.90 & -- \\
Hy4 Preview~\citep{tencent2026hy4} & -- & -- & -- & 66.40 & -- \\
Hy3~\citep{tencent2026hy3} & 84.20 & 21.30 & 74.80 & 54.90 & -- \\
DeepSeek-V4-Pro~\citep{deepseek2026v4} & 83.40 & 13.30 & \underline{76.00} & 62.00 & -- \\
\midrule
\multicolumn{6}{l}{\textit{DeepSeek-V4.1-Flash~\citep{deepseek2026v41flash}}} \\
FrontierAgent ReAct~\citep{yao2023react,apodex2026apodex11} & 73.93 & 25.00 & 74.00 & 65.23 & 36.13 \\
FrontierAgent Agent Team~\citep{apodex2026apodex11} & 87.28 & \underline{40.00} & 75.00 & 68.75 & 35.81 \\
DeepSeek Harness~\citep{deepseek-harness2026} & 80.33 & 31.67 & \underline{76.00} & 66.80 & 39.89 \\
HTN-Orchestrate (\textbf{Ours}) & 89.73 & \textbf{43.33} & \textbf{79.00} & \underline{73.44} & \textbf{46.00} \\
\bottomrule
\end{tabular}
\end{table*}

\section{Empirical Evaluations}
\label{sec:evaluation}
\subsection{Experimental Setup}
\label{sec:experimental-setup}
We evaluate HTN-Orchestrate on five benchmarks covering web research, scientific reasoning, and research workflows. BrowseComp (BC)~\citep{wei2025browsecomp} tests an agent's ability to locate hard-to-find information through persistent web browsing. For scientific reasoning and research, we use both tracks of FrontierScience~\citep{wang2026frontierscience}, with FrontierScience-Research (FSR) comprising open-ended research subproblems and FrontierScience-Olympiad (FSO) assessing olympiad-level reasoning in physics, chemistry, and biology. We use the text-only subset of 256 questions from SUPERChem~\citep{zhao2025superchem} to further evaluate multistep reasoning across chemistry subfields, while PRL-Bench~\citep{miao2026prlbench} extends the evaluation to theoretical and computational physics research tasks derived from published papers. We report accuracy for BC, FSO, and SUPERChem, where answers can be directly verified against ground truth. For the rubric-based benchmarks, we report the success rate for FSR, defined as the fraction of responses earning at least seven of ten rubric points, and the mean percentage of satisfied rubric criteria on a 0--100 scale for PRL-Bench.

We compare HTN-Orchestrate with the ReAct and Agent Team workflows implemented in FrontierAgent~\citep{apodex2026apodex11}.\footnote{\url{https://github.com/ApodexAI/FrontierAgent}} The ReAct baseline~\citep{yao2023react} uses a single agent that alternates reasoning and tool use, incorporating observations into subsequent actions. Agent Team uses a main agent to decompose the task, dispatch subagents in parallel, and integrate their reports as execution proceeds. We also include DeepSeek Harness (DSH)~\citep{deepseek-harness2026}, DeepSeek's official agent runtime in Minimal Mode, a single-agent configuration with a persistent shell and file editor, to which we add web search tools for BrowseComp.

All four methods use DeepSeek-V4.1-Flash as the backbone model with high reasoning effort. In HTN-Orchestrate, workers instead use low reasoning effort and operate in two complementary roles for research and verification. Their available tools include web search, webpage retrieval, scholarly search, Python execution, file reading, and report submission, though search and retrieval are disabled for FSR, FSO, SUPERChem, and PRL-Bench. To bound network growth and execution cost, we set the maximum task-network depth to two (counting the root at depth zero) with at most four children per node, and within each executable subtask allow up to three concurrent workers over at most twelve orchestration rounds per epoch, while the Planner can rebuild the task network once during execution. Appendix~\ref{app:experimental-details} lists the full execution settings, evaluation judges, and measurement definitions.

\subsection{Main Results}
\label{sec:main-results}

HTN-Orchestrate outperforms all three controlled baselines on every benchmark and is competitive with frontier proprietary models despite using DeepSeek-V4.1-Flash, an open-weight model, as its backbone (Table~\ref{tab:main-results}). On BrowseComp, it achieves 89.73\% accuracy, exceeding ReAct by 15.80 percentage points and Agent Team by 2.45 points. Only GPT-6 Astra (91.50\%), Kimi K3 (91.20\%), and GPT-5.6 Sol (90.40\%) score higher, all of which are frontier-scale models. Agent Team, which dispatches subagents without hierarchical planning, already exceeds ReAct by 13.35 points, confirming that multi-agent execution is broadly beneficial for persistent web research. The additional 2.45-point gain from HTN-Orchestrate over Agent Team reflects the contribution of structured decomposition and result-informed planning.

On the two open-ended research benchmarks, HTN-Orchestrate achieves the highest scores among all models in Table~\ref{tab:main-results}. On FSR, it reaches a success rate of 43.33\%, surpassing Meta Muse Spark (38.00\%), GPT-5.5 (33.90\%), and Agent Team (40.00\%). The 18.33-point margin over ReAct and 3.33-point margin over Agent Team suggest that separating global task planning from local worker coordination allows the framework to manage the iterative hypothesis refinement that open-ended research demands. PRL-Bench reinforces this pattern, with a mean rubric score of 46.00 exceeding the next-best result from Gemini 3.1 Pro (44.27) by 1.73 points, while the 10.19-point gain over Agent Team is the largest on any benchmark.

\begin{figure}[t]
\centering
\begin{subfigure}[t]{0.495\linewidth}
\centering
\includegraphics[width=\linewidth]{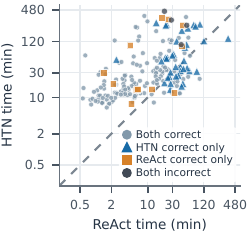}
\caption{HTN vs.\ ReAct.}
\end{subfigure}\hspace{0.005\linewidth}%
\begin{subfigure}[t]{0.495\linewidth}
\centering
\includegraphics[width=\linewidth]{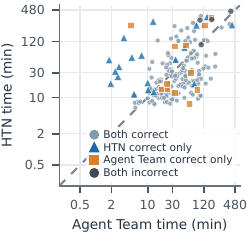}
\caption{HTN vs.\ Agent Team.}
\end{subfigure}
\caption{Per-question completion times on a 200-question subset of BrowseComp, with marker shapes and colors indicating paired correctness. Both axes use logarithmic scales; points below the dashed equality line indicate shorter completion times for HTN.}
\label{fig:bc-paired-time}
\end{figure}

On the two accuracy-based science benchmarks, HTN-Orchestrate remains competitive with much larger proprietary systems. On FSO, it reaches 79.00\% accuracy, matching Gemini 3.1 Pro and improving over Agent Team (75.00\%) and DeepSeek Harness (76.00\%) by 4.00 and 3.00 points. On SUPERChem, it scores 73.44\%, falling only 0.16 points short of GPT-5.6 Sol (73.60\%), the overall best, while exceeding Agent Team and ReAct by 4.69 and 8.21 points. Across all five benchmarks, gains over Agent Team range from 2.45 points on BC to 10.19 points on PRL-Bench, with the largest improvements appearing on tasks that reward structured decomposition and iterative refinement.

These results also highlight the complementary strengths of multi-agent orchestration and minimal-tooling runtimes. DeepSeek Harness, as a single-agent runtime, outperforms Agent Team on PRL-Bench (39.89 vs.\ 35.81) and matches it on FSO (76.00 vs.\ 75.00), showing that a focused single-agent runtime can be competitive on benchmarks that reward sustained reasoning. HTN-Orchestrate outperforms both approaches across all five benchmarks. That it achieves these results with DeepSeek-V4.1-Flash, an open-weight model, while matching or exceeding larger proprietary systems underscores the importance of orchestration design in determining overall system performance.

\subsection{Analysis}
\label{sec:analysis}

To understand how HTN-Orchestrate's two-layer design translates accuracy gains into distinct execution patterns, and to isolate the contribution of each architectural component, we conduct execution analysis and ablation studies on a 200-question subset randomly sampled from BrowseComp. The execution analysis examines completion time, worker concurrency, and internal task-network statistics to reveal the operational signatures of hierarchical orchestration, while the ablations disable result-informed planning, local multi-agent orchestration, and task-network rebuilding in turn to quantify their individual effects.

\begin{figure}[t]
\centering
\begin{subfigure}[t]{0.495\linewidth}
\centering
\includegraphics[width=\linewidth]{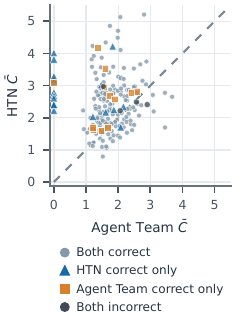}
\caption{Average concurrency $\bar C$.}
\end{subfigure}\hspace{0.005\linewidth}%
\begin{subfigure}[t]{0.495\linewidth}
\centering
\includegraphics[width=\linewidth]{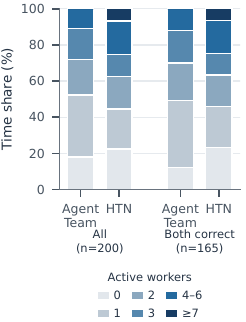}
\caption{Concurrency time shares.}
\end{subfigure}
\caption{Worker concurrency on a 200-question subset of BrowseComp. (a) Average worker concurrency $\bar C$ over each question's full trace window; the dashed line marks equality. (b) Per-question time shares, averaged equally over all 200 questions or the 165 questions both methods answer correctly. Workers are active from invocation to completion or cancellation, including model and tool waits; control agents are excluded.}
\label{fig:bc-worker-concurrency}
\end{figure}

\begin{table}[t]
\caption{HTN-Orchestrate's internal statistics on a 200-question subset of BrowseComp, grouped by outcome. Tree metrics describe the final task network; execution counts include work before and after rebuilding. Values are means except for the percentage of questions with a rebuild.}
\label{tab:bc-htn-internal}
\centering
\small
\setlength{\tabcolsep}{3pt}
\begin{tabular}{@{}lrr@{}}
\toprule
Metric & Correct & Incorrect \\
 & ($n=184$) & ($n=16$) \\
\midrule
Final tree maximum depth & 1.23 & 1.19 \\
Final tree leaves & 3.95 & 4.00 \\
Executed leaf instances & 4.52 & 6.31 \\
Worker invocations & 21.04 & 47.50 \\
Orchestrator decision rounds & 20.21 & 44.62 \\
Task-network rebuild (\%) & 14.7 & 50.0 \\
\bottomrule
\end{tabular}
\end{table}

\begin{table}[t]
\caption{Ablation results on a 200-question subset of BrowseComp. $\Delta$ is the accuracy difference from HTN-Orchestrate in percentage points.}
\label{tab:bc-ablation}
\centering
\small
\setlength{\tabcolsep}{4pt}
\begin{tabular}{@{}lrr@{}}
\toprule
Variant & Accuracy (\%) & $\Delta$ \\
\midrule
HTN-Orchestrate & \textbf{92.0} & --- \\
Upfront decomposition & 90.5 & $-1.5$ \\
Single-agent leaf execution & 89.5 & $-2.5$ \\
No rebuilding & 89.5 & $-2.5$ \\
\bottomrule
\end{tabular}
\end{table}

\begin{figure}[t]
\centering
\captionsetup[subfigure]{skip=2pt}
\begin{subfigure}[t]{0.495\linewidth}
\centering
\includegraphics[width=\linewidth]{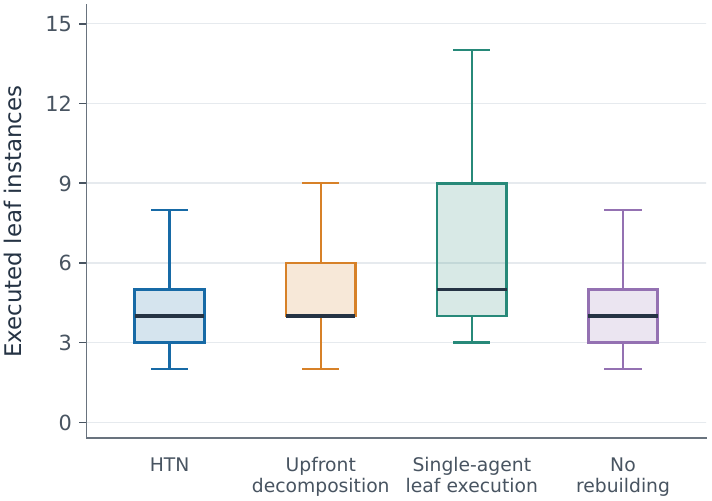}
\caption{Executed leaf instances.}
\end{subfigure}\hspace{0.005\linewidth}%
\begin{subfigure}[t]{0.495\linewidth}
\centering
\includegraphics[width=\linewidth]{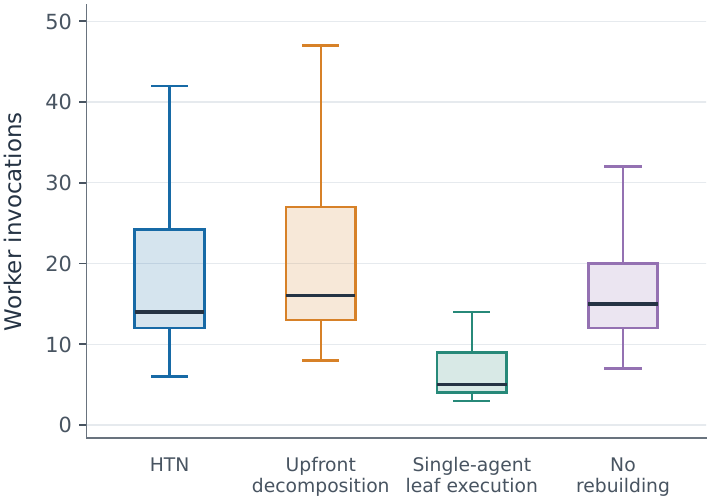}
\caption{Worker invocations.}
\end{subfigure}

\smallskip
\begin{subfigure}[t]{0.495\linewidth}
\centering
\includegraphics[width=\linewidth]{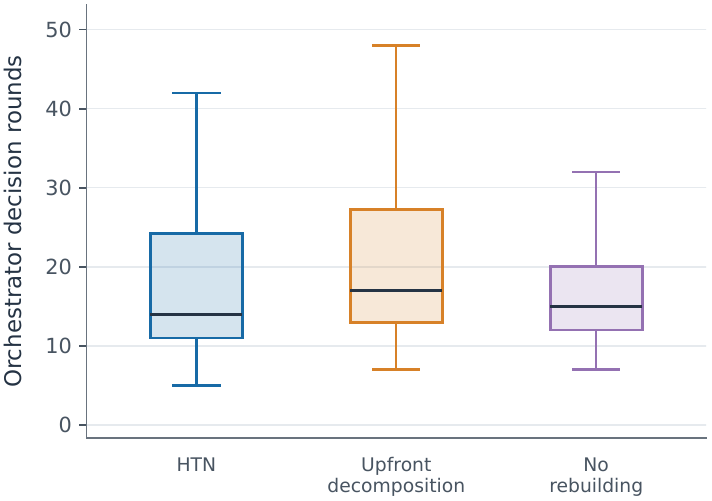}
\caption{Orchestrator decision rounds.}
\end{subfigure}\hspace{0.005\linewidth}%
\begin{subfigure}[t]{0.495\linewidth}
\centering
\includegraphics[width=\linewidth]{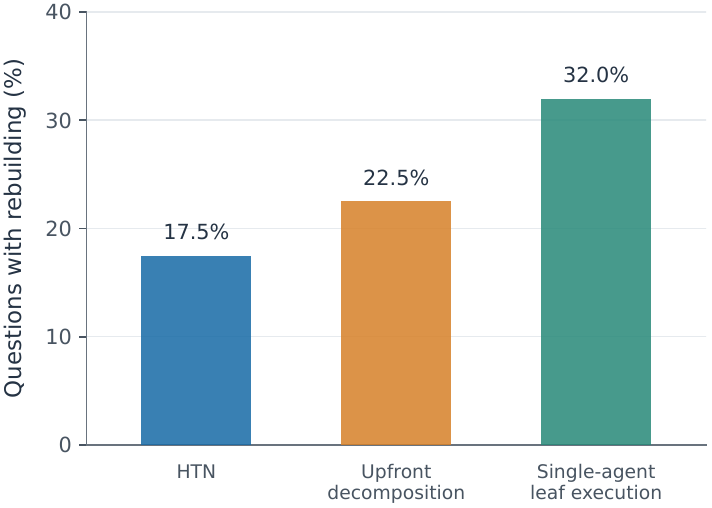}
\caption{Task-network rebuilding.}
\end{subfigure}
\caption{Execution statistics on a 200-question subset of BrowseComp for HTN-Orchestrate (HTN) and its ablated variants. Panels (a--c) show per-question counts across all epochs. Boxes span the interquartile range, horizontal lines mark medians, and whiskers extend to 1.5 times the interquartile range; outlier markers are hidden. Panel (d) shows the percentage of questions with a task-network rebuild. Single-agent leaf execution is omitted from (c), and No rebuilding from (d), because their values are uniformly zero.}
\label{fig:bc-ablation-activity}
\end{figure}

\begin{figure*}[t]
\centering
\captionsetup[subfigure]{skip=2pt}
\begin{subfigure}[t]{0.495\textwidth}
\centering
\includegraphics[width=\linewidth]{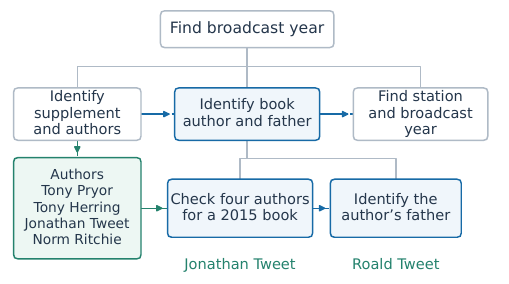}
\caption{HTN-Orchestrate.}
\end{subfigure}\hspace{0.005\textwidth}%
\begin{subfigure}[t]{0.495\textwidth}
\centering
\includegraphics[width=\linewidth]{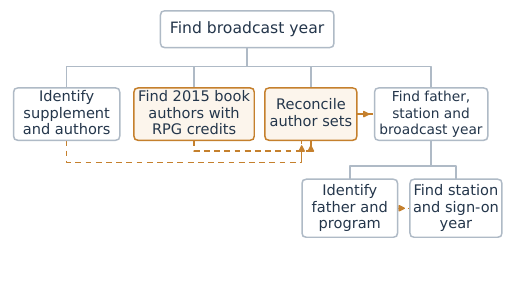}
\caption{Upfront decomposition.}
\end{subfigure}
\caption{Result-informed planning on BrowseComp Q193. The question traces an author of a 1993 role-playing-game supplement through a 2015 children's book and his father's radio program to a station's first broadcast year. Solid lines show task hierarchy; dashed arrows show dependencies or returned evidence. Green boxes contain research findings. In (a), the blue subtree is expanded after the author list is returned. In (b), the orange tasks independently discover and reconcile author candidates within a tree fully planned before research begins. Both runs correctly identify WVIK and answer 1980.}
\label{fig:case-result-informed}
\end{figure*}

\begin{figure*}[t]
\centering
\includegraphics[width=\textwidth]{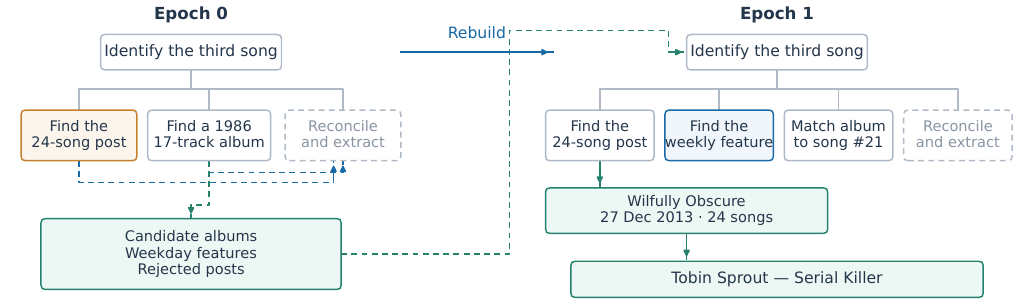}
\caption{Task-network rebuilding on BrowseComp Q501. The question asks for the third song in a 24-song blog list dated December 27, using a weekly-feature clue and the 21st song's album as additional constraints. The orange node marks the initial incomplete post search; the blue node marks the added weekly-feature task. Green boxes show retained findings and subsequent results. Solid lines show hierarchy, and dashed arrows show selected dependencies and evidence transfers. Gray dashed nodes were planned but not executed; the final answer is supported by the completed post-search branch in epoch 1.}
\label{fig:case-replanning}
\end{figure*}

\textbf{Execution Analysis.} The per-question comparisons in Figure~\ref{fig:bc-paired-time} reveal how HTN-Orchestrate's accuracy advantage interacts with completion time. Among the 165 questions both HTN-Orchestrate and Agent Team answer correctly, HTN-Orchestrate finishes faster on 112 (67.9\%), with a median per-question completion-time ratio of 0.699 relative to Agent Team, so its time advantage complements its higher overall accuracy. The comparison with ReAct reveals a different tradeoff, as HTN-Orchestrate generally takes longer on questions both methods answer correctly while correctly answering 38 questions that ReAct misses and missing only 12 that ReAct answers correctly, for a net gain of 26 correct answers.

To characterize parallel execution, we apply the average-concurrency measure used in performance analysis~\citep{huang2010anatomizing} to worker invocations. For each question, average worker concurrency is $\bar C = \frac{1}{T}\sum_i d_i$, where $d_i$ is the active duration of worker invocation $i$ within the trace window and $T$ is the question's makespan, giving the time-weighted mean number of active workers over the full window including periods with no active workers. Figure~\ref{fig:bc-worker-concurrency} compares this measure and the fraction of time spent at different worker counts. Averaged across all questions, worker concurrency is 2.47 for HTN-Orchestrate and 1.70 for Agent Team; on the 165 shared successes, it remains higher at 2.42 vs.\ 1.82. On this shared subset, HTN-Orchestrate spends an average of 24.6\% of each run with at least four active workers, compared with 11.8\% for Agent Team, but it also spends a larger fraction of time with no active workers, during which planning, coordination, or synthesis can still proceed. This pattern of concentrated worker activity interspersed with non-worker periods is consistent with its hierarchical organization, which dispatches workers concurrently across executable subtasks, and accompanies the shorter completion times observed on shared successes.

Within HTN-Orchestrate, Table~\ref{tab:bc-htn-internal} relates final task structure to the execution activity accumulated across the initial and rebuilt task networks. The two outcome groups have similar final tree depths and leaf counts, while questions scored incorrect involve more leaf executions, worker invocations, and Orchestrator decisions on average. Rebuilding also occurs more frequently in this group, at 50.0\% compared with 14.7\% for questions scored correct. Similar final task structures therefore correspond to substantially different execution histories, making accumulated orchestration and rebuilding activity important for understanding the research process.

The traces also document the use of dynamic decomposition and recovery. The Planner produces child subtasks at nodes with prerequisite dependencies 79 times across 57 of the 200 questions, confirming that result-informed planning occurs regularly during execution. Task-network rebuilding occurs on 35 questions, of which 27 are ultimately scored correct. These runs show that research can continue through a revised task network and reach a correct answer after the original execution encounters a recovery condition.

\textbf{Ablation Studies.} To isolate the contribution of each component, we evaluate three ablated variants, removing result-informed planning, local multi-agent orchestration, and task-network rebuilding in turn while preserving the remaining mechanisms (Appendix~\ref{app:additional-results} gives detailed definitions and extended results). All three reduce accuracy (Table~\ref{tab:bc-ablation}), but with distinct effects on execution behavior (Figure~\ref{fig:bc-ablation-activity}; counts cover all epochs, with No rebuilding restricted to its single epoch).

Upfront decomposition is the most direct test of the central mechanism because it changes only whether decomposition occurs before or after prerequisite results return, while retaining dependency-based execution, local multi-agent orchestration, and rebuilding. Without current-epoch research results to narrow subsequent decomposition, accuracy decreases from 92.0\% to 90.5\%. This less targeted planning also increases coordination effort at the median, with worker invocations rising from 14 to 16, Orchestrator rounds from 14 to 17, and the rebuilding rate from 17.5\% to 22.5\%. Median completion time also rises from 25.23 to 30.02 minutes. The full system therefore achieves higher accuracy with less coordination and lower latency, indicating that intermediate findings guide more effective task decomposition.

Single-agent leaf execution replaces each leaf's local orchestration with a single ReAct researcher while retaining HTN planning, synthesis, and task-network rebuilding. Accuracy drops to 89.5\%, and although the variant naturally uses fewer workers and makes no Orchestrator decisions, the mean number of executed leaf instances increases from 4.67 to 6.15 per question while the rebuilding rate nearly doubles from 17.5\% to 32.0\%. Less effective research within individual leaves shifts more of the burden to the Planner, which compensates by spawning additional subtasks and rebuilding more frequently. The full system combines higher accuracy with fewer leaf executions and less frequent rebuilding, highlighting how local orchestration supports global task resolution.

No rebuilding disables task-network reconstruction while preserving result-informed planning within the initial epoch. Accuracy decreases to 89.5\%, and because the system can no longer trigger a second epoch, average execution activity also falls, with executed leaf instances dropping from 4.67 to 4.12 per question and worker invocations from 23.16 to 19.01. The full system's higher accuracy therefore comes at the cost of this additional research, which the rebuilding mechanism makes possible by retaining findings and revising the task network when the initial decomposition cannot resolve the task.

\textbf{Case Studies.} Figure~\ref{fig:case-result-informed} shows how intermediate findings shape task decomposition on BrowseComp question Q193. After identifying the supplement and its four authors, HTN-Orchestrate incorporates their names into a newly generated task that checks which author wrote a children's book in 2015. The upfront decomposition variant instead specifies an independent search for children's-book authors with role-playing-game credits, followed by a separate task to reconcile the two candidate sets. Both runs reach the correct answer, but the result-informed plan narrows the downstream search to the returned author list and executes four leaves rather than five.

Figure~\ref{fig:case-replanning} illustrates task-network rebuilding on Q501. The initial search does not locate the target music-blog post. HTN-Orchestrate retains candidate albums, recurring weekday features, and rejected posts, then rebuilds the network with an additional feature-based search and an explicit album-to-list matching task. Research continues until the post-search branch locates the target article and verifies its song list. The final answer is \textit{Serial Killer} by Tobin Sprout, while No rebuilding returns \textit{Silent Night}. Appendix~\ref{app:case-studies} presents an additional case study covering targeted verification within a leaf.

\section{Conclusion}
\label{sec:conclusion}
We presented HTN-Orchestrate, a two-layer orchestration framework that separates global task planning from local multi-agent execution. Its upper layer maintains a hierarchical task network that decomposes incrementally as prerequisite findings become available and can be rebuilt when the initial plan proves insufficient, while its lower layer coordinates multiple workers within each executable subtask. On five benchmarks spanning web research, scientific reasoning, and research workflows, HTN-Orchestrate consistently outperforms ReAct, Agent Team, and DeepSeek Harness. Ablation studies on BrowseComp suggest that result-informed planning, local multi-agent orchestration, and task-network rebuilding each contribute to this advantage, with execution traces revealing interdependence between the two layers. These findings show that separating control across two layers while feeding results back through a shared, revisable task structure allows each layer to inform the other and strengthens both. Future work includes scaling the task-network depth for problems that require deeper decomposition hierarchies and evaluating the framework across additional backbone models.

\section*{Acknowledgments}
This work was supported by Apodex. Yanchen Deng, Jianguo Huang, Yao Long Teng, and Yixuan Li conducted this research during internships at Apodex.

\bibliography{example_paper}

@inproceedings{yang2023leandojo,
  author    = {Yang, Kaiyu and Swope, Aidan and Gu, Alex and Chalamala, Rahul and Song, Peiyang and Yu, Shixing and Godil, Saad and Prenger, Ryan J and Anandkumar, Animashree},
  title     = {{LeanDojo}: {T}heorem Proving with Retrieval-Augmented Language Models},
  booktitle = {NeurIPS},
  pages     = {21573--21612},
  year      = {2023}
}

@article{novikov2025alphaevolve,
  title={{AlphaEvolve}: {A} coding agent for scientific and algorithmic discovery},
  author={Novikov, Alexander and V{\~u}, Ng{\^a}n and Eisenberger, Marvin and Dupont, Emilien and Huang, Po-Sen and Wagner, Adam Zsolt and Shirobokov, Sergey and Kozlovskii, Borislav and Ruiz, Francisco JR and Mehrabian, Abbas and others},
  journal={arXiv preprint arXiv:2506.13131},
  year={2025}
}

@article{hubert2026olympiad,
  author    = {Hubert, Thomas and Mehta, Rishi and Sartran, Laurent and others},
  title     = {Olympiad-level formal mathematical reasoning with reinforcement learning},
  journal   = {Nature},
  volume    = {651},
  pages     = {607--613},
  year      = {2026}
}

@misc{openclaw2026,
  author       = {Steinberger, Peter and {OpenClaw Community}},
  title        = {{OpenClaw}: Your Own Personal {AI} Assistant},
  year         = {2026},
  howpublished = {\url{https://github.com/openclaw/openclaw}}
}

@misc{hermesagent2026,
  author       = {{Nous Research}},
  title        = {Hermes Agent: The Agent That Grows with You},
  year         = {2026},
  howpublished = {\url{https://github.com/NousResearch/hermes-agent}}
}

@misc{claudecode2025,
  author       = {{Anthropic}},
  title        = {Claude Code},
  year         = {2025},
  howpublished = {\url{https://www.anthropic.com/claude-code}}
}

@misc{claudemanagedagents2026,
  author       = {{Anthropic}},
  title        = {Claude Managed Agents},
  year         = {2026},
  howpublished = {\url{https://platform.claude.com/docs/en/managed-agents/overview}}
}

@inproceedings{yao2023react,
  title = {{ReAct}: Synergizing Reasoning and Acting in Language Models},
  author = {Yao, Shunyu and Zhao, Jeffrey and Yu, Dian and Du, Nan and Shafran, Izhak and Narasimhan, Karthik and Cao, Yuan},
  booktitle = {ICLR},
  year = {2023}
}

@inproceedings{wu2024autogen,
  title = {{AutoGen}: Enabling Next-Gen {LLM} Applications via Multi-Agent Conversations},
  author = {Wu, Qingyun and Bansal, Gagan and Zhang, Jieyu and Wu, Yiran and Li, Beibin and Zhu, Erkang and Jiang, Li and Zhang, Xiaoyun and Zhang, Shaokun and Liu, Jiale and Awadallah, Ahmed and White, Ryen W. and Burger, Doug and Wang, Chi},
  booktitle = {COLM},
  year = {2024}
}

@inproceedings{hong2024metagpt,
  title = {{MetaGPT}: Meta Programming for a Multi-Agent Collaborative Framework},
  author = {Hong, Sirui and Zhuge, Mingchen and Chen, Jonathan and Zheng, Xiawu and Cheng, Yuheng and Wang, Jinlin and Zhang, Ceyao and Wang, Zili and Yau, Steven and Lin, Zijuan and Zhou, Liyang and Ran, Chenyu and Xiao, Lingfeng and Wu, Chenglin and Schmidhuber, J{\"u}rgen},
  booktitle = {ICLR},
  year = {2024}
}

@article{fourney2024magentic,
  title = {{Magentic-One}: A Generalist Multi-Agent System for Solving Complex Tasks},
  author = {Fourney, Adam and Bansal, Gagan and Mozannar, Hussein and Tan, Cheng and Salinas, Eduardo and Zhu, Erkang and Niedtner, Friederike and Proebsting, Grace and Bassman, Griffin and Gerrits, Jack and Alber, Jacob and Chang, Peter and Loynd, Ricky and West, Robert and Dibia, Victor and Awadallah, Ahmed and Kamar, Ece and Hosn, Rafah and Amershi, Saleema},
  journal = {arXiv preprint arXiv:2411.04468},
  year = {2024}
}

@article{kimi2026k25,
  title = {{Kimi K2.5}: Visual Agentic Intelligence},
  author = {{Kimi Team}},
  journal = {arXiv preprint arXiv:2602.02276},
  year = {2026}
}

@article{apodex2026apodex11,
  title = {{Apodex 1.1}: Scaling Agentic Intelligence for Complex Work},
  author = {{Apodex Team}},
  journal = {arXiv preprint arXiv:2608.23283},
  year = {2026}
}

@misc{langgraph2026workflows,
  title = {Workflows and Agents},
  author = {{LangChain}},
  howpublished = {{LangGraph} documentation},
  year = {2026},
  note = {Accessed September 17, 2026},
  url = {https://docs.langchain.com/oss/python/langgraph/workflows-agents}
}

@inproceedings{prasad2024adapt,
  title = {{ADaPT}: As-Needed Decomposition and Planning with Language Models},
  author = {Prasad, Archiki and Koller, Alexander and Hartmann, Mareike and Clark, Peter and Sabharwal, Ashish and Bansal, Mohit and Khot, Tushar},
  booktitle = {Findings of NAACL},
  pages = {4226--4252},
  year = {2024}
}

@inproceedings{choi2026reactree,
  title = {{ReAcTree}: Hierarchical {LLM} Agent Trees with Control Flow for Long-Horizon Task Planning},
  author = {Choi, Jae-Woo and Kim, Hyungmin and Ong, Hyobin and Yoon, Youngwoo and Jang, Minsu and Kim, Dohyung and Kim, Jaehong},
  booktitle = {AAMAS},
  year = {2026}
}

@inproceedings{erol1994htn,
  title = {{HTN} Planning: Complexity and Expressivity},
  author = {Erol, Kutluhan and Hendler, James and Nau, Dana S.},
  booktitle = {AAAI},
  pages = {1123--1128},
  year = {1994}
}

@inproceedings{zhang2025aflow,
  title = {{AFlow}: Automating Agentic Workflow Generation},
  author = {Zhang, Jiayi and Xiang, Jinyu and Yu, Zhaoyang and Teng, Fengwei and Chen, Xionghui and Chen, Jiaqi and Zhuge, Mingchen and Cheng, Xin and Hong, Sirui and Wang, Jinlin and Zheng, Bingnan and Liu, Bang and Luo, Yuyu and Wu, Chenglin},
  booktitle = {ICLR},
  year = {2025}
}

@article{ruan2026aorchestra,
  title = {{AOrchestra}: Automating Sub-Agent Creation for Agentic Orchestration},
  author = {Ruan, Jianhao and Xu, Zhihao and Peng, Yiran and Ren, Fashen and Yu, Zhaoyang and Liang, Xinbing and Xiang, Jinyu and Chen, Yongru and Liu, Bang and Wu, Chenglin and Luo, Yuyu and Zhang, Jiayi},
  journal = {arXiv preprint arXiv:2602.03786},
  year = {2026}
}

@article{zhang2025agentorchestra,
  title = {{AgentOrchestra}: Orchestrating Multi-Agent Intelligence with the {Tool-Environment-Agent (TEA)} Protocol},
  author = {Zhang, Wentao and Zeng, Liang and Xiao, Yuzhen and Li, Yongcong and Cui, Ce and Zhao, Yilei and Hu, Rui and Liu, Yang and Zhou, Yahui and An, Bo},
  journal = {arXiv preprint arXiv:2506.12508},
  year = {2025}
}

@inproceedings{wang2025openhands,
  title = {{OpenHands}: An Open Platform for {AI} Software Developers as Generalist Agents},
  author = {Wang, Xingyao and Li, Boxuan and Song, Yufan and Xu, Frank F. and Tang, Xiangru and Zhuge, Mingchen and Pan, Jiayi and Song, Yueqi and Li, Bowen and Singh, Jaskirat and Tran, Hoang and Li, Fuqiang and Ma, Ren and Zheng, Mingzhang and Qian, Bill and Shao, Daniel and Muennighoff, Niklas and Zhang, Yizhe and Hui, Binyuan and Lin, Junyang and Brennan, Robert and Peng, Hao and Ji, Heng and Neubig, Graham},
  booktitle = {ICLR},
  year = {2025}
}

@inproceedings{agashe2025agents2,
  title = {{Agent S2}: A Compositional Generalist-Specialist Framework for Computer Use Agents},
  author = {Agashe, Saaket and Wong, Kyle and Tu, Vincent and Yang, Jiachen and Li, Ang and Wang, Xin Eric},
  booktitle = {COLM},
  year = {2025}
}

@inproceedings{zhang2025maas,
  title = {Multi-agent Architecture Search via Agentic Supernet},
  author = {Zhang, Guibin and Niu, Luyang and Fang, Junfeng and Wang, Kun and Bai, Lei and Wang, Xiang},
  booktitle = {ICML},
  pages = {75834--75852},
  year = {2025}
}

@inproceedings{nielsen2026conductor,
  title = {Learning to Orchestrate Agents in Natural Language with the Conductor},
  author = {Nielsen, Stefan and Cetin, Edoardo and Schwendeman, Peter and Sun, Qi and Xu, Jinglue and Tang, Yujin},
  booktitle = {ICLR},
  year = {2026}
}

@inproceedings{ke2026masorchestra,
  title = {{MAS-Orchestra}: Understanding and Improving Multi-Agent Reasoning Through Holistic Orchestration and Controlled Benchmarks},
  author = {Ke, Zixuan and Ming, Yifei and Xu, Austin and Chin, Ryan and Nguyen, Xuan-Phi and Jwalapuram, Prathyusha and Yavuz, Semih and Xiong, Caiming and Joty, Shafiq},
  booktitle = {ICML},
  year = {2026}
}

@inproceedings{dang2025evolving,
  title = {Multi-Agent Collaboration via Evolving Orchestration},
  author = {Dang, Yufan and Qian, Chen and Luo, Xueheng and Fan, Jingru and Xie, Zihao and Shi, Ruijie and Chen, Weize and Yang, Cheng and Che, Xiaoyin and Tian, Ye and Xiong, Xuantang and Han, Lei and Liu, Zhiyuan and Sun, Maosong},
  booktitle = {NeurIPS},
  pages = {165025--165059},
  year = {2025}
}

@inproceedings{su2026toolorchestra,
  title = {{ToolOrchestra}: Elevating Intelligence via Efficient Model and Tool Orchestration},
  author = {Su, Hongjin and Diao, Shizhe and Lu, Ximing and Liu, Mingjie and Xu, Jiacheng and Dong, Xin and Fu, Yonggan and Belcak, Peter and Ye, Hanrong and Yin, Hongxu and Dong, Yi and Bakhturina, Evelina and Yu, Tao and Choi, Yejin and Kautz, Jan and Molchanov, Pavlo},
  booktitle = {ICML},
  year = {2026}
}

@inproceedings{hu2025owl,
  title = {{OWL}: Optimized Workforce Learning for General Multi-Agent Assistance in Real-World Task Automation},
  author = {Hu, Mengkang and Zhou, Yuhang and Fan, Wendong and Nie, Yuzhou and Xia, Bowei and Sun, Tao and Ye, Ziyu and Jin, Zhaoxuan and Li, Yingru and Chen, Qiguang and Zhang, Zeyu and Wang, Yifeng and Ye, Qianshuo and Ghanem, Bernard and Luo, Ping and Li, Guohao},
  booktitle = {NeurIPS},
  pages = {50859--50906},
  year = {2025}
}

@inproceedings{zhou2023leasttomost,
  title = {Least-to-Most Prompting Enables Complex Reasoning in Large Language Models},
  author = {Zhou, Denny and Sch{\"a}rli, Nathanael and Hou, Le and Wei, Jason and Scales, Nathan and Wang, Xuezhi and Schuurmans, Dale and Cui, Claire and Bousquet, Olivier and Le, Quoc and Chi, Ed},
  booktitle = {ICLR},
  year = {2023}
}

@inproceedings{li2025aop,
  title = {Agent-Oriented Planning in Multi-Agent Systems},
  author = {Li, Ao and Xie, Yuexiang and Li, Songze and Tsung, Fugee and Ding, Bolin and Li, Yaliang},
  booktitle = {ICLR},
  year = {2025}
}

@inproceedings{erdogan2025planandact,
  title = {Plan-and-Act: Improving Planning of Agents for Long-Horizon Tasks},
  author = {Erdogan, Lutfi Eren and Lee, Nicholas and Kim, Sehoon and Moon, Suhong and Furuta, Hiroki and Anumanchipalli, Gopala and Keutzer, Kurt and Gholami, Amir},
  booktitle = {ICML},
  pages = {15419--15462},
  year = {2025}
}

@inproceedings{liu2025selectdecompose,
  title = {Select-Then-Decompose: From Empirical Analysis to Adaptive Selection Strategy for Task Decomposition in Large Language Models},
  author = {Liu, Shuodi and Liu, Yingzhuo and Wang, Zi and Wang, Yusheng and Wu, Huijia and Xiang, Liuyu and He, Zhaofeng},
  booktitle = {EMNLP},
  pages = {5454--5477},
  year = {2025}
}

@inproceedings{xu2026dcore,
  title = {{D-CORE}: Incentivizing Task Decomposition in Large Reasoning Models for Complex Tool Use},
  author = {Xu, Bowen and Wu, Shaoyu and Jiang, Hao and Liu, Kai and Chen, Xin and Hu, Lulu and Yang, Bin},
  booktitle = {ICML},
  year = {2026}
}

@inproceedings{zhang2025recap,
  title = {{ReCAP}: Recursive Context-Aware Reasoning and Planning for Large Language Model Agents},
  author = {Zhang, Zhenyu and Chen, Tianyi and Xu, Weiran and Pentland, Alex and Pei, Jiaxin},
  booktitle = {NeurIPS},
  pages = {95158--95186},
  year = {2025}
}

@inproceedings{liu2026todoevolve,
  title = {{TodoEvolve}: Learning to Architect Agent Planning Systems},
  author = {Liu, Jiaxi and Jiang, Yanzuo and Zhang, Guibin and Zhang, Zihan and Chang, Heng and Yin, Zhenfei and Ren, Qibing and Yan, Junchi},
  booktitle = {ICML},
  year = {2026}
}

@inproceedings{munozavila2025chathtn,
  title = {{ChatHTN}: Interleaving Approximate ({LLM}) and Symbolic {HTN} Planning},
  author = {Mu{\~n}oz-Avila, H{\'e}ctor and Aha, David W. and Rizzo, Paola},
  booktitle = {International Conference on Neuro-symbolic Systems},
  pages = {446--458},
  year = {2025}
}

@inproceedings{kim2024llmcompiler,
  title = {An {LLM} Compiler for Parallel Function Calling},
  author = {Kim, Sehoon and Moon, Suhong and Tabrizi, Ryan and Lee, Nicholas and Mahoney, Michael W. and Keutzer, Kurt and Gholami, Amir},
  booktitle = {ICML},
  pages = {24370--24391},
  year = {2024}
}

@article{hu2025flowsearch,
  title = {{FlowSearch}: Advancing Deep Research with Dynamic Structured Knowledge Flow},
  author = {Hu, Yusong and Ma, Runmin and Fan, Yue and Shi, Jinxin and Cao, Zongsheng and Zhou, Yuhao and Yuan, Jiakang and Zhang, Shuaiyu and Feng, Shiyang and Yan, Xiangchao and Zhang, Shufei and Zhang, Wenlong and Bai, Lei and Zhang, Bo},
  journal = {arXiv preprint arXiv:2510.08521},
  year = {2025}
}

@article{wei2025browsecomp,
  title = {{BrowseComp}: A Simple Yet Challenging Benchmark for Browsing Agents},
  author = {Wei, Jason and Sun, Zhiqing and Papay, Spencer and McKinney, Scott and Han, Jeffrey and Fulford, Isa and Chung, Hyung Won and Passos, Alex Tachard and Fedus, William and Glaese, Amelia},
  journal = {arXiv preprint arXiv:2504.12516},
  year = {2025}
}

@article{wang2026frontierscience,
  title = {{FrontierScience}: Evaluating {AI}'s Ability to Perform Expert-Level Scientific Tasks},
  author = {Wang, Miles and Lin, Robi and Hu, Kat and Jiao, Joy and Chowdhury, Neil and Chang, Ethan and Patwardhan, Tejal},
  journal = {arXiv preprint arXiv:2601.21165},
  year = {2026}
}

@article{zhao2025superchem,
  title = {{SUPERChem}: A Multimodal Reasoning Benchmark in Chemistry},
  author = {Zhao, Zehua and Huang, Zhixian and Li, Junren and Lin, Siyu and Zhou, Junting and Cao, Fengqi and Zhou, Kun and Ge, Rui and Long, Tingting and Zhu, Yuexiang and Liu, Yan and Zheng, Jie and Wei, Junnian and Zhu, Rong and Zou, Peng and Li, Wenyu and Cheng, Zekai and Ding, Tian and Wang, Yaxuan and Yan, Yizhao and Wei, Tingru and Ming, Haowei and Mao, Weijie and Sun, Chen and Liu, Yiming and Wang, Zichen and Zhang, Zuo and Yang, Tong and Ma, Hao and Gao, Zhen and Pei, Jian},
  journal = {arXiv preprint arXiv:2512.01274},
  year = {2025}
}

@article{miao2026prlbench,
  title = {{PRL-Bench}: A Comprehensive Benchmark Evaluating {LLMs}' Capabilities in Frontier Physics Research},
  author = {Miao, Tingjia and Jin, Wenkai and Zhang, Muhua and Tan, Jinxin and Hu, Yuelin and Guo, Tu and Zhang, Jiejun and Wang, Yuhan and Li, Wenbo and Gao, Yinuo and Chen, Shuo and Jiang, Weiqi and Hu, Yayun and Lei, Zixing and Pang, Xianghe and Liu, Zexi and Zhang, Yuzhi and Zhang, Linfeng and Chen, Kun and Wang, Wei and E, Weinan and Chen, Siheng},
  journal = {arXiv preprint arXiv:2604.15411},
  year = {2026}
}

@inproceedings{huang2010anatomizing,
  title = {Anatomizing Application Performance Differences on Smartphones},
  author = {Huang, Junxian and Xu, Qiang and Tiwana, Birjodh and Mao, Z. Morley and Zhang, Ming and Bahl, Paramvir},
  booktitle = {MobiSys},
  pages = {165--178},
  year = {2010},
  doi = {10.1145/1814433.1814452}
}

@misc{deepseek-harness2026,
  title={DeepSeek Harness: Everything is a Plugin},
  author={{DeepSeek-AI}},
  year={2026},
  publisher={GitHub},
  howpublished={\url{https://github.com/deepseek-ai/deepseek-harness}},
}

@misc{openai2025gpt5,
  title = {{GPT-5.4} Thinking},
  author = {{OpenAI}},
  year = {2026},
  howpublished = {\url{https://deploymentsafety.openai.com/gpt-5-4-thinking}}
}

@misc{openai2026gpt55,
  title = {{GPT-5.5} System Card},
  author = {{OpenAI}},
  year = {2026},
  howpublished = {\url{https://openai.com/index/gpt-5-5-system-card}}
}

@misc{openai2026gpt56,
  title = {Introducing {GPT-5.6}},
  author = {{OpenAI}},
  year = {2026},
  howpublished = {\url{https://openai.com/index/gpt-5-6/}}
}

@misc{openai2026gpt6astra,
  title = {{GPT-6} Astra},
  author = {{OpenAI}},
  year = {2026},
  howpublished = {\url{https://openai.com/blog/gpt-6-astra}}
}

@misc{anthropic2026opus46,
  title = {Claude {Opus} 4.6},
  author = {{Anthropic}},
  year = {2026},
  howpublished = {\url{https://www.anthropic.com/news/claude-opus-4-6}}
}

@misc{anthropic2026opus48,
  title = {Claude {Opus} 4.8},
  author = {{Anthropic}},
  year = {2026},
  howpublished = {\url{https://www.anthropic.com/research/claude-opus-4-8}}
}

@misc{anthropic2026fable5,
  title = {Claude {Fable} 5},
  author = {{Anthropic}},
  year = {2026},
  howpublished = {\url{https://www.anthropic.com/research/claude-fable-5}}
}

@misc{google2026gemini31,
  title = {Gemini 3.1 {Pro} Model Card},
  author = {{Google DeepMind}},
  year = {2026},
  howpublished = {\url{https://deepmind.google/models/model-cards/gemini-3-1-pro/}}
}

@misc{qwen2026qwen37,
  title = {Qwen3.7: The Agent Frontier},
  author = {{Qwen Team}},
  year = {2026},
  howpublished = {\url{https://qwen.ai/blog?id=qwen3.7}}
}

@misc{meta2026musespark,
  title = {Muse Spark},
  author = {{Meta Superintelligence Labs}},
  year = {2026},
  howpublished = {\url{https://ai.meta.com/blog/muse-spark/}}
}

@article{kimi2026k3,
  title = {{Kimi K3}: Open Frontier Intelligence},
  author = {{Kimi Team}},
  journal = {arXiv preprint arXiv:2607.24653},
  year = {2026}
}

@misc{tencent2026hy3,
  title = {Hunyuan {Hy3}},
  author = {{Tencent}},
  year = {2026},
  howpublished = {\url{https://huggingface.co/tencent/Hy3}}
}

@misc{tencent2026hy4,
  title = {Hunyuan {Hy4} Preview},
  author = {{Tencent}},
  year = {2026},
  howpublished = {\url{https://huggingface.co/tencent/Hy4-Preview}}
}

@article{deepseek2026v4,
  title = {{DeepSeek-V4}: Towards Highly Efficient Million-Token Context Intelligence},
  author = {{DeepSeek-AI}},
  journal = {arXiv preprint arXiv:2606.19348},
  year = {2026}
}

@misc{deepseek2026v41flash,
  title = {{DeepSeek-V4.1-Flash} Technical Report},
  author = {{DeepSeek-AI}},
  year = {2026},
  howpublished = {\url{https://huggingface.co/deepseek-ai/DeepSeek-V4.1-Flash}}
}
\bibliographystyle{icml2026}

\newpage
\onecolumn

\appendix

\section{Experimental Details}
\label{app:experimental-details}

\subsection{Benchmarks and Tool Access}

We evaluate HTN-Orchestrate on five benchmarks covering web research, scientific problem solving, and research-level reasoning. Table~\ref{tab:app-benchmarks} summarizes the evaluation sets. The BrowseComp main results use the full 1{,}266-question set, and the ablations use the same fixed subset of 200 questions for all variants.

\begin{table}[h]
\caption{Evaluation sets and solver tools. Python execution is available in all five settings. Offline evaluation disables search, page retrieval, and external network access while retaining Python computation.}
\label{tab:app-benchmarks}
\centering
\small
\begin{tabular}{@{}lrlc@{}}
\toprule
Benchmark & Questions & Task format & Retrieval \\
\midrule
BrowseComp & 1{,}266 & Multi-hop web research; short factual answers & Enabled \\
BC ablation subset & 200 & Same task format; fixed question IDs across variants & Enabled \\
FrontierScience-Research & 60 & Research-level scientific problems; written solutions & Disabled \\
FrontierScience-Olympiad & 100 & Olympiad-level problems; worked solutions and final answers & Disabled \\
SUPERChem & 256 & Text-only chemistry questions; option selection & Disabled \\
PRL-Bench & 100 & Research-level physics problems; written solutions & Disabled \\
\bottomrule
\end{tabular}
\end{table}

For offline tasks, worker instructions ask agents to derive results from the problem statement, check assumptions, and verify calculations with Python. The verifier checks the derivation and answer independently using the same available tools.

\subsection{Model and Execution Settings}

The solver backbone is DeepSeek-V4.1-Flash. Table~\ref{tab:app-exec-settings} gives the execution settings used across all benchmarks. Planner and Orchestrator calls use high reasoning effort, while worker and instruction-generation calls use low reasoning effort. Leaf synthesis, tree synthesis, and recovery use high reasoning effort.

\begin{table}[h]
\caption{HTN-Orchestrate execution settings. Tree depth counts the root as depth zero. Worker capacity applies within a leaf; independent ready leaves may execute concurrently.}
\label{tab:app-exec-settings}
\centering
\small
\begin{tabular}{@{}lr@{}}
\toprule
Setting & Value \\
\midrule
Temperature / top-$p$ & 1.0 / 0.95 \\
Maximum output tokens per model call & 65{,}536 \\
Maximum tree depth & 2 \\
Maximum children per decomposition & 4 \\
Maximum Orchestrator decisions per leaf per epoch & 12 \\
Worker capacity per leaf & 3 \\
Maximum ReAct turns per worker & 70 \\
Maximum tree rebuilds per question & 1 \\
Leaf time budget & 7{,}200\,s \\
Question time budget & 28{,}800\,s \\
Finalization reserve & 1{,}800\,s \\
Completed precursor context budget & 24{,}000 chars total; 6{,}000 per summary \\
Repeated runs per question and variant & 1 \\
\bottomrule
\end{tabular}
\end{table}

The ReAct, Agent Team, and DeepSeek Harness baselines use the same solver backbone in the within-backbone comparison. Their results are reported separately from the published model results in the main table.

\subsection{Evaluation Judges}

BrowseComp, FrontierScience-Research, FrontierScience-Olympiad, and PRL-Bench use GPT-5.5 as the judge for semantic answer matching and rubric evaluation. SUPERChem uses exact matching against the selected option.

\subsection{Execution Measurements}

Question duration is the elapsed time recorded for the selected question attempt, including planning, worker execution, coordination, and finalization. Execution counts include all epochs of that attempt. An executed leaf is identified by its question, epoch, and node ID, and a worker invocation is one ReAct worker launch. Orchestrator decisions include dispatch, wait, and terminate actions.

For worker concurrency, let $n_q(t)$ be the number of active workers at time $t$ during question $q$, and let $T_q$ be the duration of its main execution trace. The time-averaged concurrency $\bar C$ used in Section~\ref{sec:analysis} is
\begin{equation}
\bar C_q = \frac{1}{T_q}\int_0^{T_q} n_q(t)\,dt.
\end{equation}
This measure includes intervals spent planning or synthesizing when no worker is active. Dataset averages give each question equal weight. Paired latency ratios compare the same question only when both methods answer it correctly.

\section{Implementation Details}
\label{app:implementation}

\subsection{Task-Node Representation}

Each task node stores its objective, hierarchical position, sibling prerequisites, execution state, and evidence summary. Table~\ref{tab:app-node-fields} lists the fields used by planning and execution. A pending leaf becomes ready when all its declared prerequisites have completed. If further planning is required, the Planner receives the node objective and completed precursor summaries before choosing between direct execution and decomposition.

\begin{table}[h]
\caption{Task-node representation.}
\label{tab:app-node-fields}
\centering
\small
\begin{tabular}{@{}ll@{}}
\toprule
Field & Meaning \\
\midrule
\texttt{id}, \texttt{parent\_id}, \texttt{children} & Node identity and hierarchical edges \\
\texttt{title}, \texttt{task} & Short label and self-contained objective \\
\texttt{depends\_on} & Sibling nodes whose completion is required before this node becomes ready \\
\texttt{needs\_decomposition} & Whether a ready node returns to the Planner \\
\texttt{status} & \texttt{pending}, \texttt{running}, \texttt{completed}, or \texttt{failed} \\
\texttt{result\_summary}, \texttt{ref} & Evidence summary and reference to the full leaf result \\
\texttt{completion\_outcome} & \texttt{resolved} or \texttt{exhausted\_unresolved} \\
\texttt{failure\_reason} & Execution or synthesis failure information \\
\bottomrule
\end{tabular}
\end{table}

\subsection{Prompt Templates and Tool Schemas}

Table~\ref{tab:app-role-io} summarizes the information supplied to each role and the output it controls. The remainder of this section excerpts the core decision logic from each role's system prompt and lists the tool schemas that structure their outputs. Full prompt templates are available in the code repository.

\begin{table}[h]
\caption{Role inputs and outputs.}
\label{tab:app-role-io}
\centering
\small
\begin{tabular}{@{}lp{7.5cm}p{5.5cm}@{}}
\toprule
Role & Input & Output \\
\midrule
Planner & Root question, current task, completed precursor summaries, restart context & Direct execution or validated child tasks and dependencies \\
Orchestrator & Leaf objective, prerequisite evidence, returned reports, active assignments, budget & Dispatch, wait, or terminate \\
Instruction generator & Leaf objective, selected role, evidence and remaining issue & An executable worker assignment \\
Researcher & Assigned objective, original constraints, prior evidence & Findings, evidence, unresolved constraints, confidence \\
Verifier & Proposed answer, original objective, supporting reports & Completeness and correctness checks, verdict \\
Synthesizer & Node objective and worker or child reports & Evidence summary and completion outcome \\
Recovery & Original question, retained evidence, unresolved relations & A revised task network \\
\bottomrule
\end{tabular}
\end{table}

\paragraph{Planner.}
The Planner receives the root question, the current task to decompose, completed precursor summaries, and (when rebuilding) retained evidence from a previous epoch. Its user message follows this structure:

\noindent\fbox{\parbox{\dimexpr\textwidth-2\fboxsep-2\fboxrule}{%
\small\ttfamily
\#\# Root Goal\\
\{root\_question\}\\[4pt]
\#\# Prior Epoch Recovery Context \textnormal{\textit{[present only when rebuilding]}}\\
\{retained\_evidence\}\\[4pt]
\#\# Task to Decompose\\
\{current\_task\_title\}\\
\{current\_task\_objective\}\\[4pt]
\#\# Completed Precursor Evidence \textnormal{\textit{[present when prerequisites have completed]}}\\
<precursor relation="direct|ancestor" node\_id="\{id\}"\\
\hspace*{2em}title="\{title\}" completion\_outcome="\{outcome\}">\\
\{completed\_result\_summary\}\\
</precursor>\\[4pt]
\#\# Decomposition Constraints\\
- Maximum subtasks: \{max\_children\}\\
- Depth remaining after this split: \{child\_remaining\_depth\}
}}

\medskip
\noindent The system prompt instructs the Planner to choose exactly one planning action:

\noindent\fbox{\parbox{\dimexpr\textwidth-2\fboxsep-2\fboxrule}{%
\small\ttfamily
Choose exactly one planning action:\\
- execute\_direct when the remaining work is one coherent investigation,
  verification, or synthesis that the adaptive leaf orchestrator can handle.\\
- decompose only when two or more genuinely distinct workstreams benefit
  from separate execution, parallelism, timeout isolation, or later
  hierarchical synthesis.
}}

\medskip
\noindent The Planner emits its decision through tool calls. For decomposition, it emits one \texttt{decompose} call per subtask. The runtime collects these calls into a single atomic plan, validates coverage and acyclicity, and installs the children.

\noindent\fbox{\parbox{\dimexpr\textwidth-2\fboxsep-2\fboxrule}{%
\small\ttfamily
decompose(id, title, task, depends\_on, needs\_decomposition, rationale)\\
\hspace*{1em}id:\hspace*{3.5em}short unique slug for this subtask\\
\hspace*{1em}title:\hspace*{2.5em}human-readable label (3--7 words)\\
\hspace*{1em}task:\hspace*{3em}compact structural objective ({$\leq$}1000 chars)\\
\hspace*{1em}depends\_on: IDs of sibling subtasks that must finish first (default: [])\\
\hspace*{1em}needs\_decomposition: whether to return to planning after prerequisites\\
\hspace*{1em}rationale:\hspace*{0.5em}why this branch is warranted ({$\leq$}800 chars)
}}

\medskip
\noindent For direct execution, a single \texttt{execute\_direct(rationale)} call marks the current node as an executable leaf.

\paragraph{Orchestrator.}
The Orchestrator receives the complete leaf context and all returned reports at every decision round. The excerpt below shows the retrieval-enabled rendering used for BrowseComp. When retrieval is disabled, the template conditionally omits search and retrieval references from role descriptions and dispatch instructions. The user message contains:

\noindent\fbox{\parbox{\dimexpr\textwidth-2\fboxsep-2\fboxrule}{%
\small\ttfamily
\#\# Task Context (C)\\
\{leaf\_objective\_and\_prerequisite\_evidence\}\\[4pt]
\#\# Sub-agent Results (R)\\
\{completed\_worker\_reports\}\\[4pt]
\#\# In-flight Agent Work (do not duplicate)\\
\{outstanding\_assignments\}\\[4pt]
\#\# Available Agent Roles\\
dynir\_researcher -- searches, fetches, computes; gathers facts + evidence\\
dynir\_verifier\hspace*{1.1em}-- independently re-checks a candidate answer\\[4pt]
\#\# Execution Budget\\
Rounds remaining: \{rounds\_remaining\}\\
Max agents this round: \{available\_worker\_capacity\}\\[4pt]
Call one of: terminate, dispatch, wait.
}}

\medskip
\noindent The core decision rule from the system prompt:

\noindent\fbox{\parbox{\dimexpr\textwidth-2\fboxsep-2\fboxrule}{%
\small\ttfamily
dispatch -- Send agents when R does not yet fully + verifiably answer the
leaf task. Pursue important routes through genuinely different retrieval
methods. Verify BEFORE finishing: once you have a candidate, dispatch a
verifier so it re-confirms with independent searches.\\[4pt]
wait -- Use when R is not sufficient to terminate, at least one agent is
still in flight, and R exposes no distinct gap worth dispatching.\\[4pt]
terminate -- When R contains a complete answer AND a verifier has confirmed
it, OR when no distinct evidence-producing route remains. Retain the
best-supported candidate and explicit unresolved gaps.
}}

\medskip
\noindent Each \texttt{dispatch} call specifies \texttt{role\_id}, \texttt{task\_summary}, and \texttt{rationale}. Multiple \texttt{dispatch} calls in one response start independent assignments concurrently.

\paragraph{Researcher.}
The Researcher receives a self-contained assignment produced by the instruction generator. The system prompt specifies a five-step methodology (understand, search strategically, extract with precision, compute and verify in code, write a complete answer) and requires all calculations to use Python. The report is submitted through a \texttt{submit\_report(content, confidence)} tool call:

\noindent\fbox{\parbox{\dimexpr\textwidth-2\fboxsep-2\fboxrule}{%
\small\ttfamily
Scope: [what was investigated]\\
Finding: [best-supported finding or ranked candidates]\\
Constraint ledger:\\
\hspace*{1em}[candidate] -- SUPPORTED / CONTRADICTED / UNRESOLVED for each constraint\\
Evidence:\\
\hspace*{1em}Source URL or computation\\
Confidence: [numeric value from 0.0 to 1.0]\\
Conflicts: [contradictions between sources with citations, or "none"]
}}

\paragraph{Verifier.}
The Verifier receives the proposed answer, the original task, and supporting reports. Its system prompt specifies four phases: (1) completeness, listing every requirement as answered or missing; (2) correctness, independently verifying factual claims and recomputing calculations; (3) precision, confirming exact values rather than approximations; and (4) answer contract, checking scope and format. Its report uses:

\noindent\fbox{\parbox{\dimexpr\textwidth-2\fboxsep-2\fboxrule}{%
\small\ttfamily
Verdict: CONFIRMED | NEEDS CORRECTION | INCOMPLETE\\
Unconfirmed or wrong constraints: [specific factual gaps, or "none"]\\
Corrected answer: [supported replacement, or "N/A"]
}}

\paragraph{Synthesis.}
Separate templates cover leaf, internal-node, and root synthesis. Leaf synthesis joins worker reports into an evidence memo. Internal synthesis combines child results for the parent. Root synthesis checks every requested output field and returns the final answer. All templates distinguish \texttt{resolved} (the complete answer-bearing claim is positively supported) from \texttt{exhausted\_unresolved} (a decisive relation remains missing after available routes are exhausted, with partial evidence preserved).

\paragraph{Recovery.}
When the current task network cannot resolve the root objective, the recovery step extracts supported facts through an \texttt{extract\_supported\_facts} tool that requires each claim to include a verbatim supporting quote and source reference. These facts, together with unresolved relations and the reason the previous approach stopped, form the restart context for the next epoch. The rebuild instructions supplement the Planner with the requirement to identify the most discriminative unresolved relation and to preserve a discovery route that can surface an alternative candidate.

\section{Additional Results and Ablation Details}
\label{app:additional-results}

\subsection{Ablation Definitions}

All three ablations use the same 200 BrowseComp questions. Each changes one part of the orchestration procedure while preserving the remaining mechanisms.

\begin{table}[h]
\caption{Components changed by each ablated variant.}
\label{tab:app-ablation-defs}
\centering
\small
\begin{tabular}{@{}p{3.2cm}p{6.5cm}p{5.5cm}@{}}
\toprule
Variant & Changed behavior & Retained components \\
\midrule
Upfront decomposition & Complete the task network before executing any leaf in each epoch. Decomposition instructions refer to anticipated outputs rather than completed precursor results. & Dependency-constrained execution, local multi-agent orchestration, synthesis, rebuilding \\
Single-agent leaf execution & Launch exactly one ReAct researcher for each executable leaf. Its report goes directly to leaf synthesis, replacing Orchestrator decisions and multi-worker dispatch. & Result-informed planning, prerequisite context, leaf and tree synthesis, rebuilding \\
No rebuilding & Set the maximum number of rebuilds to zero. & Result-informed planning, local multi-agent orchestration, synthesis, finalization from accumulated evidence \\
\bottomrule
\end{tabular}
\end{table}

In the single-agent variant, the unit of assignment is an executed leaf within an epoch. A question may still contain several leaves, and a rebuilt task network may introduce new leaves. In the upfront variant, a downstream worker can consume completed prerequisite results even though its position and objective were planned earlier.

\subsection{Accuracy, Execution Volume, and Latency}

Table~\ref{tab:app-ablation-results} extends the accuracy comparison in the main paper with execution volume and latency measurements. Upfront decomposition uses more executed leaves and workers than the full system and has longer mean and median duration. The single-agent variant reduces worker use and latency while its number of executed leaves increases and rebuilding occurs on 64 questions, compared with 35 for the full system. Removing rebuilding reduces mean duration, although its median duration is slightly higher.

\begin{table}[h]
\caption{BrowseComp-200 ablation results. Execution counts and duration include all epochs of each selected attempt. Time is measured in minutes; leaves, workers, and decisions are means per question.}
\label{tab:app-ablation-results}
\centering
\small
\setlength{\tabcolsep}{4pt}
\begin{tabular}{@{}lcccccccr@{}}
\toprule
Method & Correct & Acc.\ (\%) & Mean time & Med.\ time & Exec.\ leaves & Workers & Decisions & Rebuilt \\
\midrule
HTN-Orchestrate & 184 & 92.0 & 58.03 & 25.23 & 4.665 & 23.155 & 22.165 & 35 \\
Upfront decomposition & 181 & 90.5 & 67.69 & 30.02 & 5.025 & 24.760 & 24.025 & 45 \\
Single-agent leaf execution & 179 & 89.5 & 28.49 & 17.69 & 6.145 & 6.145 & 0 & 64 \\
No rebuilding & 179 & 89.5 & 46.54 & 28.15 & 4.120 & 19.010 & 18.500 & 0 \\
\bottomrule
\end{tabular}
\end{table}

\subsection{Paired Correctness}

\begin{figure}[h]
\centering
\includegraphics[width=0.6\linewidth]{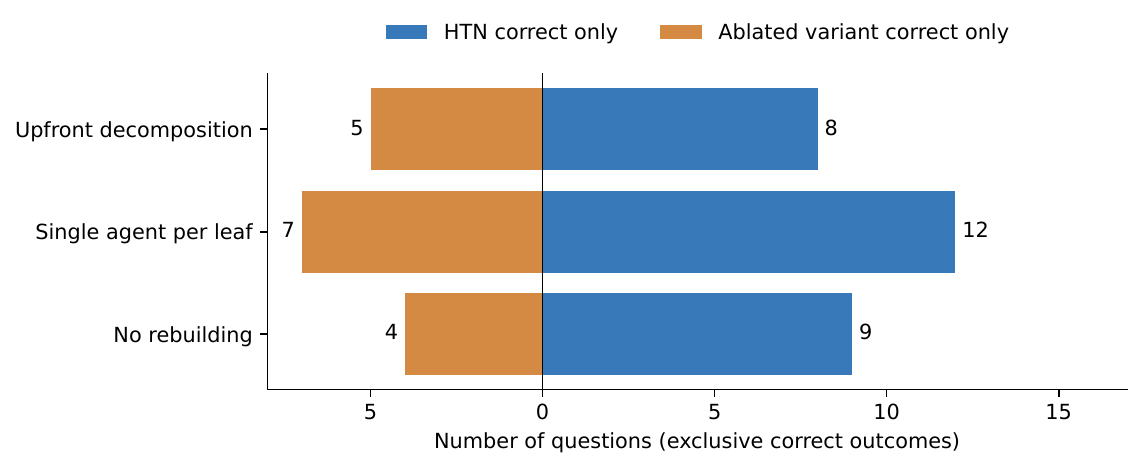}
\caption{Exclusive correct outcomes in the ablation comparison. Blue bars count questions answered correctly only by HTN-Orchestrate; orange bars count questions answered correctly only by the ablated variant. The difference gives the net gain in Table~\ref{tab:app-paired}.}
\label{fig:app-ablation-paired}
\end{figure}

Table~\ref{tab:app-paired} partitions the same 200 questions by paired correctness between the full system and each ablated variant. Figure~\ref{fig:app-ablation-paired} visualizes the exclusive correct outcomes.

\begin{table}[h]
\caption{Paired correctness on BrowseComp-200. ``HTN correct only'' means the full system is correct and the ablated variant is incorrect; ``Ablated variant correct only'' denotes the reverse.}
\label{tab:app-paired}
\centering
\small
\begin{tabular}{@{}lccccr@{}}
\toprule
Ablated variant & Both correct & HTN correct only & Ablated variant correct only & Both incorrect & Net HTN gain \\
\midrule
Upfront decomposition & 176 & 8 & 5 & 11 & $+3$ \\
Single-agent leaf execution & 172 & 12 & 7 & 9 & $+5$ \\
No rebuilding & 175 & 9 & 4 & 12 & $+5$ \\
\bottomrule
\end{tabular}
\end{table}

\section{Additional Case Studies}
\label{app:case-studies}

\subsection{Using an Intermediate Result to Specialize Later Work}

BrowseComp Q193 asks for a radio station's first broadcast year through a chain of clues involving a 1993 role-playing supplement, one of its authors, a children's book, and the author's father. The initial investigation identifies \textit{Creative Campaigning} and returns four authors: Tony Pryor, Tony Herring, Jonathan Tweet, and Norm Ritchie. The Planner then specializes the downstream investigation using this returned list, checking which author wrote the 2015 children's book and establishing the father's identity.

The resulting chain identifies Jonathan Tweet, \textit{Grandmother Fish}, Roald Tweet, and WVIK, leading to the answer \textbf{1980}. The upfront variant also answers correctly, but plans its children's-book investigation before receiving the author list. Across the complete attempts, HTN-Orchestrate executes four leaves and 15 workers, compared with five leaves and 20 workers for upfront decomposition. This case shows how an intermediate result can make the remaining task more specific even when both methods eventually reach the same answer.

\subsection{Targeted Verification within a Leaf}

BrowseComp Q1025 asks for a conductor identified by a 2008 concert review, a poem-based opening work, and a venue dedicated in the early 1990s. The task network separates identifying the poem, locating candidate reviews, and reconciling the conductor's identity. Within the review-search leaf, the Orchestrator dispatches complementary review-first, venue-first, and opening-song investigations.

The venue-first report connects the Berger Center, dedicated in 1992, with Reveille Men's Chorus, conductor \textbf{Linus Lerner}, and the opener \textit{Come Travel With Me}. The Orchestrator then assigns a verifier to check that evidence chain. The completed attempt uses three leaves and ten workers. The single-agent-per-leaf variant executes nine leaves across two epochs and submits \textbf{Gregory Wait}; its final explanation does not establish the specific 2008 review connecting that conductor to the requested performance.

\begin{figure}[h]
\centering
\includegraphics[width=\linewidth]{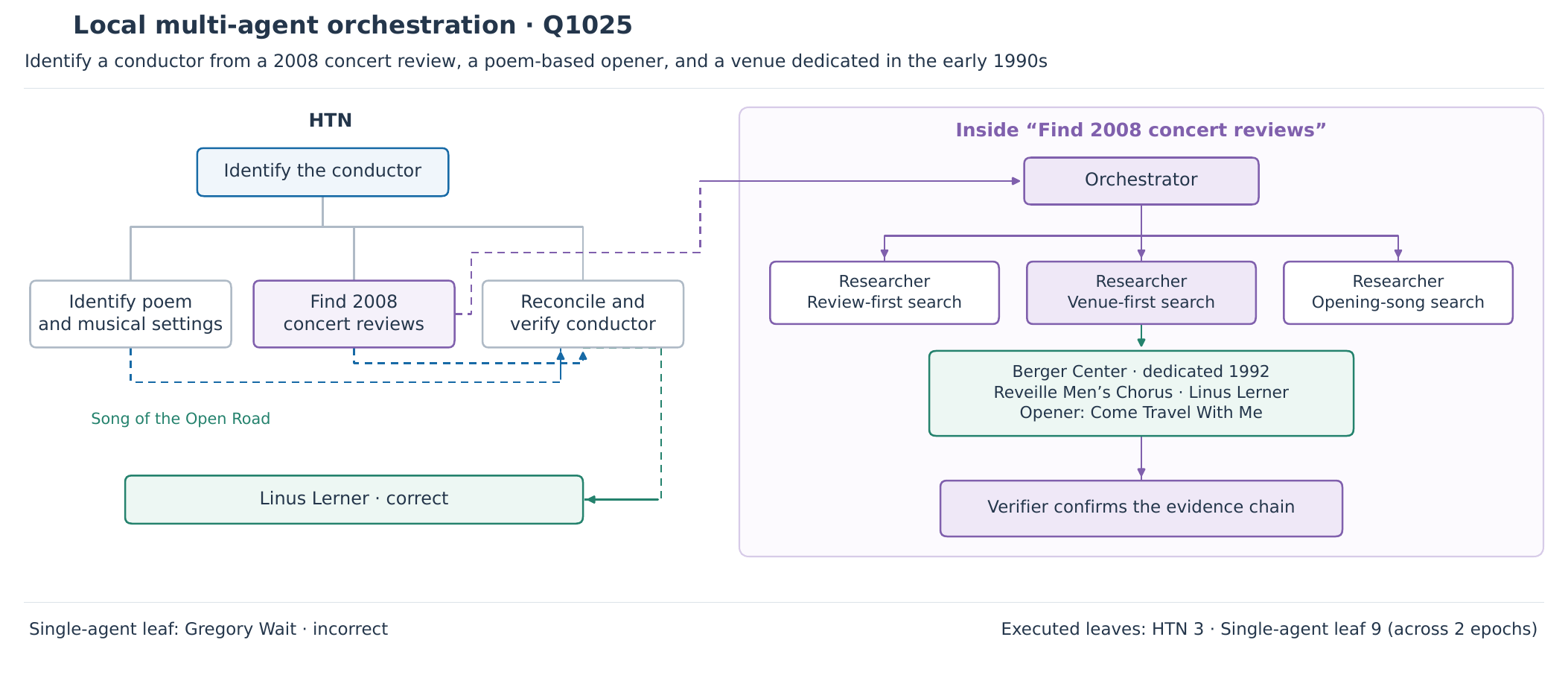}
\caption{Local multi-agent orchestration on BrowseComp Q1025. The expanded leaf shows complementary discovery assignments followed by verification of a returned candidate. The full system identifies Linus Lerner; the single-agent-per-leaf variant submits Gregory Wait.}
\label{fig:app-local-orch}
\end{figure}

\subsection{Evidence Carried across a Rebuild}
\label{app:case-rebuild}

BrowseComp Q501 asks for the third song in a 24-song blog list dated 27 December, using clues about a weekly feature and an album. In the first epoch, the investigation fails to recover the required list before a leaf timeout. Recovery retains the findings and unresolved source-identification work, and the next epoch constructs a revised task network.

The second epoch obtains answer-bearing evidence in its post-list branch. When a later timeout ends further execution, recovery finalization uses the accumulated evidence to return \textbf{``Serial Killer'' by Tobin Sprout}. The no-rebuilding variant returns \textbf{``Silent Night.''}\ The full attempt executes five leaves and 72 workers across two epochs, compared with two leaves and 18 workers for no rebuilding. This case illustrates a successful continuation through evidence-preserving recovery, including finalization from the evidence available at the execution limit.

\begin{table}[h]
\caption{Execution counts for the three BrowseComp case pairs. Counts cover the entire selected question attempt.}
\label{tab:app-case-exec}
\centering
\small
\begin{tabular}{@{}llrrrrl@{}}
\toprule
Question & Method & Leaves & Workers & Decisions & Rebuilds & Answer \\
\midrule
Q193 & HTN-Orchestrate & 4 & 15 & 18 & 0 & 1980 \\
Q193 & Upfront decomposition & 5 & 20 & 21 & 0 & 1980 \\
Q1025 & HTN-Orchestrate & 3 & 10 & 10 & 0 & Linus Lerner \\
Q1025 & Single-agent leaf execution & 9 & 9 & 0 & 1 & Gregory Wait \\
Q501 & HTN-Orchestrate & 5 & 72 & 56 & 1 & Serial Killer \\
Q501 & No rebuilding & 2 & 18 & 18 & 0 & Silent Night \\
\bottomrule
\end{tabular}
\end{table}

\end{document}